\documentclass{article}

\PassOptionsToPackage{numbers, compress}{natbib}

\usepackage[main, final]{neurips_2026}

\usepackage{amsmath}

\usepackage[utf8]{inputenc} 
\usepackage[T1]{fontenc}    
\usepackage{hyperref}       
\usepackage{url}            
\usepackage{booktabs}       
\usepackage{amsfonts}       
\usepackage{nicefrac}       
\usepackage{microtype}      
\usepackage{xcolor}         

\usepackage{wrapfig}
\usepackage{tcolorbox}
\usepackage{graphicx}
\usepackage{xspace}
\usepackage{xcolor}
\usepackage{subcaption}
\newcommand{\benchmark}{\textsc{TokenSwap-Bench}\xspace}

\definecolor{PhotoshopUIGray}{RGB}{60,60,60}
\tcbuselibrary{listings, breakable}
\newtcblisting{promptbox}[1][]{
  colback=gray!5,
  colframe=black!50,
  boxrule=0.5mm,
  arc=2mm,
  title=#1,
  listing only,
    listing options={
      basicstyle=\ttfamily\scriptsize,
      breaklines=true,
      breakindent=0pt,
      breakautoindent=false,
      columns=fullflexible,
      keepspaces=true,
  }
}

\title{TokenSwap: Benchmarking and Reducing the \\Modality Gap in Multimodal LLMs}

\author{%
\setlength{\tabcolsep}{12pt}

\begin{tabular}{cccc}
\textbf{Andong Hua$^{1}$} &
\textbf{Colton Bishop$^{2}$} &
\textbf{Igor Mordatch$^{2}$} &
\textbf{Arian Hosseini$^{2}$} \\
\textbf{Jindong Gu$^{2}$} &
\textbf{Aleksandra Faust$^{2}$} &
\textbf{Rebecca Roelofs$^{2}$} &
\textbf{Yao Qin$^{1,2}$} \\
\end{tabular}
\\[1.5em]
$^{1}$University of California, Santa Barbara \qquad
$^{2}$Google DeepMind}

\begin{document}

\maketitle

\begin{abstract}

Multimodal large language models (MLLMs) should exhibit cross-modal consistency, generating consistent responses given semantically equivalent inputs across modalities. However, we observe a systematic discrepancy in model predictions under such cross-modal variations.
Specifically, we define the modality gap as the difference in model performance under semantically equivalent textual and multimodal inputs.
We introduce TokenSwap, a method that constructs such inputs by replacing textual concepts with semantically aligned images, resulting in sequences where visual tokens are interleaved with text tokens.
Based on TokenSwap, we transform existing text-based benchmarks (e.g., MMLU~\citep{mmlu}) into image-interleaved counterparts, resulting in \benchmark.
Across 42 MLLMs, we observe a pervasive modality gap, with performance decreasing by 4.2\% to 47.4\% when moving from text-only to image-interleaved inputs, averaging 19.6\% $\pm$ 3.3\% across models.
Notably, reasoning models exhibit consistently smaller gaps, achieving an average gap of 10.1\% compared to 25.5\% for non-reasoning models, suggesting greater robustness in multimodal reasoning across equivalent textual and visual representations.
In contrast, neither prompting strategies nor scaling training compute alone reliably reduces the modality gap.
Finally, we demonstrate that incorporating TokenSwap during training effectively mitigates this gap, improving cross-modal alignment while preserving strong text-only and vision-language model performance.
The project page is available at \url{https://dongxzz.github.io/TokenSwap/}.
\end{abstract}

\begin{figure}[h] 
\centering
\includegraphics[width=0.98\textwidth]{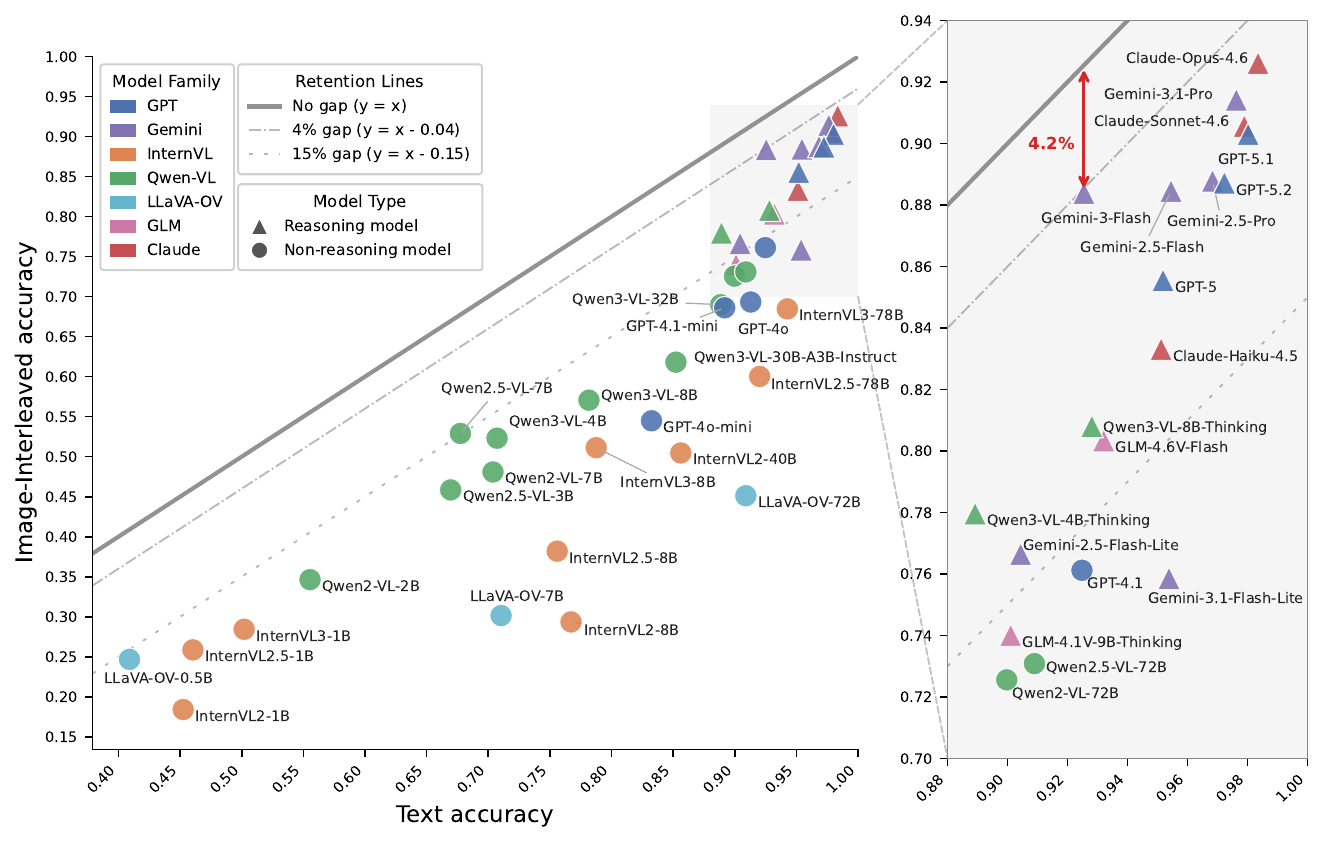}
\caption{\textbf{A pervasive modality gap exists: all models perform worse on image-interleaved inputs than on text-only inputs, with gaps ranging from 4.2\% to 47.4\%.}
Each point represents a model, colored by family, with triangles indicating reasoning models. The diagonal $y=x$ denotes equal performance, and dashed lines indicate constant gap levels (4.0\%, and 15.0\%). All models lie below the diagonal, indicating consistently lower performance under image-interleaved inputs. The right panel zooms into high-performing models. Gemini-3-Flash achieves the smallest gap (4.2\%). Reasoning models typically exhibit smaller gaps (4.0\%--15.0\%), while non-reasoning models tend to have substantially larger gaps.}
\label{fig:modality_gap}
\vspace{-3mm}
\end{figure}

\section{Introduction}

\begin{figure}[t] 
\centering
\includegraphics[width=0.98\textwidth]{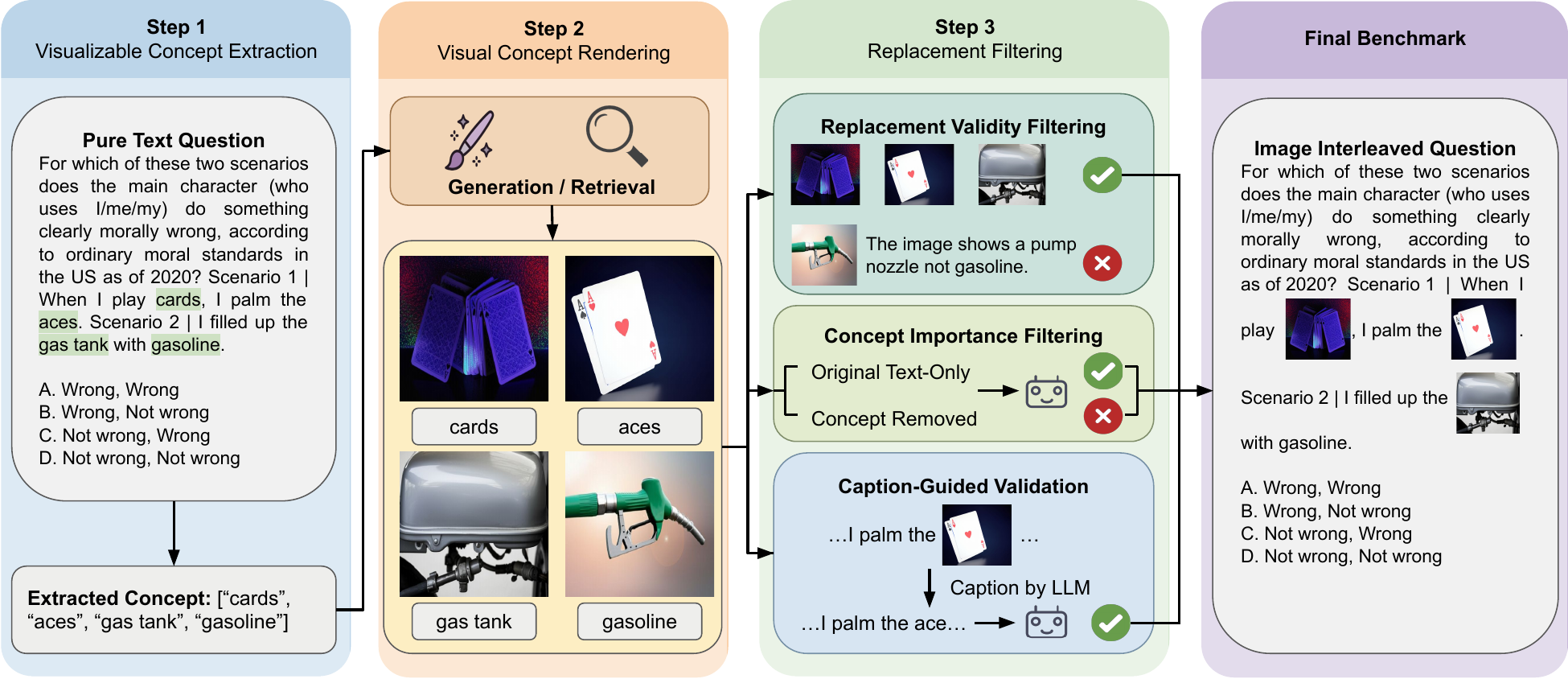}
\caption{\textbf{TokenSwap constructs semantically aligned image-interleaved questions, where Step 1 panel shows the original MMLU question and the Final Benchmark panel shows its image-interleaved counterpart.} We first extract visualizable concepts and generate or retrieve images representing them, followed by rigorous filtering to ensure validity, task relevance, and semantic consistency, resulting in semantically aligned image-interleaved questions.}
\label{fig:pipeline}
\vspace{-3mm}
\end{figure}

Multimodal Large Language Models (MLLMs) are expected to be semantically invariant, producing consistent predictions across semantically equivalent inputs regardless of whether they are presented as pure text or with interleaved images~\citep{wang2025xmodbench,tang2025seam,chen2024omnixr}. Yet, this consistency across modalities often breaks in practice. Consider the example in the Final Benchmark panel in Figure~\ref{fig:pipeline}: the image-interleaved version is semantically aligned with the original, but key concepts (e.g., ``cards'', ``aces'', ``gas tank'') have been replaced by corresponding images. Of 42 evaluated MLLMs, 11 answer the text-only version correctly but fail on the image-interleaved input, revealing a systematic violation of semantic invariance.

We formalize this phenomenon as the \emph{modality gap}, defined as the difference in model performance when semantically equivalent content is presented in textual versus multimodal form. To systematically study this gap, we introduce TokenSwap, a general method for transforming text-only datasets into image-interleaved counterparts while preserving semantic alignment across modalities. Rather than converting entire inputs into images (e.g., rendering text as images~\citep{zhang2023lost,zhang2024cross}), TokenSwap operates at the concept level, replacing individual textual concepts with semantically aligned natural images while preserving the surrounding context and structure (Figure~\ref{fig:pipeline}).

We apply TokenSwap to MMLU~\citep{mmlu} to construct \benchmark, a benchmark containing 1,516 samples with 6,946 total images, specifically designed to quantify the modality gap. Across 42 state-of-the-art MLLMs, we observe a clear modality gap: performance drops from text-only to image-interleaved inputs range from 4.2\% to 47.4\%, with an average decrease of 19.6\% ± 3.3\% (95\% confidence interval), as shown in Figure~\ref{fig:modality_gap}. Taking a closer look, we discover that reasoning-oriented models exhibit substantially smaller gaps (10.1\% on average) compared to non-reasoning models (25.5\%). 
We further find that common prompting strategies, such as chain-of-thought and few-shot prompting cannot reliably reduce the modality gap. These findings suggest that the smaller gaps of reasoning-oriented models are more likely linked to special training-time mechanisms. In addition, as shown in Figure~\ref{fig:modality_gap_size_ablation_combined}, scaling training compute yields only marginal gains, with a $10\times$ increase in FLOPs reducing the gap by approximately $2.8\%$.

Finally, unlike prior work that is primarily diagnostic~\citep{zhang2023lost,zhang2024cross,tang2025seam}, we show that augmenting text data with image-interleaved counterparts via TokenSwap effectively mitigates the modality gap in both pre-training and post-training settings. To our knowledge, this is the first training-based approach to mitigate the modality gap for multimodal LLMs. Importantly, these improvements do not come at the cost of either text-only or vision-language performance, and in some cases even lead to slight gains. Our contributions are summarized as follows:

\begin{enumerate}
    \item We propose TokenSwap, a data-centric method for constructing image-interleaved inputs by replacing textual concepts with semantically aligned images, and introduce TokenSwap-Bench, a benchmark for systematically quantifying the modality gap in MLLMs.
    \item We evaluate 42 MLLMs and find the modality gap to be pervasive. Reasoning models consistently exhibit smaller gaps, while prompting strategies and scaling alone provide limited improvement.
    \item We demonstrate that TokenSwap training effectively reduces the modality gap at both pre-training and post-training stages, while preserving strong text-only and vision-language performance.
\end{enumerate}

\section{Related Works}
\paragraph{Modality Gap and Cross-Modal Consistency.}
The notion of \emph{modality gap} has been extensively studied in contrastive vision--language models such as CLIP~\citep{clip,jia2021scaling,cherti2023reproducible,sun2023eva,zhai2023sigmoid,tschannen2025siglip}, where image and text representations may remain separated despite being embedded in a shared space, leading to representation-level discrepancies~\citep{liang2022mind,shi2023understanding,eslami2024mitigate,yaras2024explaining,fahim2024s,schrodi2024two}. 
Related work on multimodal large language models (MLLMs) instead examines \emph{cross-modal consistency}, evaluating whether models produce consistent predictions when the same semantic content is presented in different modalities~\citep{chen2024omnixr,yue2025mmmu,alonso2025vision,zhang2023lost,zhang2024cross,van2025same,tang2025seam,wang2025xmodbench}. 
These studies reveal that MLLMs often exhibit inconsistent behavior across modalities, influenced by factors such as rendering choices, visual attributes, and domain structure.
Additional details on related work are provided in Appendix~\ref{app:related_work}.

Compared with these works, our setting is distinct in two key aspects.
First, prior studies typically focus on whole-input modality conversion, rendered-text settings, or highly structured domains with standardized symbolic notations, whereas we consider an open-domain setting where natural images are interleaved into textual inputs. Notably, TokenSwap enables transforming arbitrary existing textual benchmarks into their image-interleaved counterparts. As discussed in Section~\ref{subsec:benchmark_comparison}, our benchmark provides a complementary perspective on modality gap compared to prior benchmarks. Second, most prior work is primarily diagnostic, with limited exploration of mitigation strategies. In contrast, we not only conduct a comprehensive evaluation across 42 MLLMs, including recent reasoning models, but also demonstrate that the modality gap can be effectively reduced through TokenSwap-based training.

\section{Quantifying Modality Gap: \benchmark}
\label{sec:benchmark}

To systematically study the modality gap in Multimodal Large Language Models (MLLMs), we introduce \benchmark, a benchmark designed to measure the performance discrepancy when the same semantic information is presented in different modalities. Unlike conventional vision-language benchmarks~\citep{mmbench,gqa,textvqa,mme,mmmu} that evaluate models on multimodal understanding with complementary information, our benchmark instead focuses on measuring the modality gap under semantic equivalence.

\subsection{Formalizing Modality Gap via TokenSwap}
\label{subsec:formalization}

We study modality gap under semantic equivalence: an ideal MLLM should make consistent predictions when the same content is presented in textual or visual form. Given a text input $X_{\text{text}}$, TokenSwap replaces selected visualizable concepts (e.g., words or phrases) with semantically aligned images, producing an image-interleaved input $X_{\text{interleaved}}$ while preserving the surrounding textual context, as illustrated in Figure~\ref{fig:pipeline}. We provide a more detailed token-level formalization of TokenSwap in Appendix~\ref{app:formalization}. We quantify the modality gap as
\begin{equation}
    \Delta_{\text{Gap}} =
    \text{Eval}(X_{\text{text}}) -
    \text{Eval}(X_{\text{interleaved}}),
    \label{eq:modality_gap}
\end{equation}
where $\text{Eval}(\cdot)$ denotes the model's performance metric (e.g., accuracy). 
$\Delta_{\text{Gap}} > 0$ indicates the presence of a modality gap, while larger values correspond to a more substantial modality gap.

\subsection{Benchmark Construction}
\label{subsec:pipeline}

We construct \benchmark by applying TokenSwap to MMLU~\citep{mmlu}. 
As illustrated in Figure~\ref{fig:pipeline}, the construction follows three stages. The final benchmark contains 1,516 samples with 6,946 image replacements, averaging 4.58 replacements per sample. Representative examples are shown in Figure~\ref{fig:examples}.

\subsubsection{Visualizable Concept Extraction}
We use Gemini-2.0-Flash~\citep{gemini} to identify concepts in each question and answer option that can be visually represented, typically concrete entities and visually grounded phrases. The full prompt is provided in Appendix~\ref{app:construct_prompts}.

\subsubsection{Visual Concept Rendering}

We render each visualizable concept into an image using a text-to-image model.
Specifically, given a concept $c$, we synthesize a corresponding image $I_c$ with Gemini using the prompt:
\textit{``generate an image of a $\{c\}$''}.

\subsubsection{Replacement Filtering}
To ensure that measured gaps are not driven by poor or irrelevant substitutions, we apply a rigorous sequence of filtering procedures. Across all filters, we consistently use Gemini-2.0-Flash~\citep{gemini} as the underlying LLM for filtering and as a proxy model. We provide additional implementation details in Appendix~\ref{app:filtering_details}.

\paragraph{Validity filtering:} We use an LLM to verify that the generated image faithfully represents the intended concept in context. Only replacements that preserve the original meaning are retained.

\paragraph{Importance filtering:} We ensure that the replaced concepts are task-relevant. Let $\mathcal{C}$ denote the set of valid textual concepts in a sample, and let $X_{\text{text}}^{-\mathcal{C}}$ be the input obtained by removing them from the original text. We retain only samples satisfying:
\begin{equation}
    \text{Eval}(X_{\text{text}}) = 1 \quad \text{and} \quad \text{Eval}(X_{\text{text}}^{-\mathcal{C}}) = 0.
    \label{eq:importance_filtering}
\end{equation}
This ensures that the selected concepts collectively affect the model's prediction.

\paragraph{Caption-guided validation:} We verify semantic recoverability by captioning each substituted image and replacing the visual tokens in $X_{\text{interleaved}}$ with the generated captions, yielding $\tilde{X}_{\text{text}}$. We retain only samples satisfying:
\begin{equation}
    \text{Eval}(X_{\text{text}}) = 1 \quad \text{and} \quad \text{Eval}(\tilde{X}_{\text{text}}) = 1.
    \label{eq:caption_validation}
\end{equation}
This round-trip process reduces the chance that the measured gap is caused by unrecognizable or semantically mismatched images.

We use Gemini-2.0-Flash as the underlying model for all filtering stages. To avoid directly evaluating the same model used as the proxy, we exclude Gemini-2.0-Flash from our model comparison. 
To test whether our results are sensitive to the choice of filtering proxy, we further rebuild \benchmark with GPT-5-mini as the proxy, rerunning the two stages that depend most on the proxy: importance filtering and caption-guided validation.
Across nine evaluated models, the ranking is unchanged (Spearman $\rho=1.00$, Kendall $\tau=1.00$, Pearson $r=0.99$), and Gemini-3-Flash still achieves the smallest modality gap. Full results are provided in Appendix~\ref{app:proxy_robustness}.

\subsection{Semantic Equivalence and Human Validation}
\label{subsec:semantic_equivalence}

TokenSwap relies on semantic equivalence between textual concepts and their visual counterparts. In practice, perfect equivalence is not achievable because textual concepts are abstract but images are concrete instances. We therefore use \emph{semantic recoverability} as a practical proxy: the substituted image should convey information that can be recovered as text and preserve the task prediction. In particular, validity filtering and caption-guided validation explicitly ensure this recoverability.

We further validate substitution quality through a human study. We randomly sample one question from each of the 57 subjects in \benchmark, covering 213 substituted concepts. Two annotators achieve 92.0\% (196/213) and 89.2\% (190/213) accuracy in identifying the intended concept from the image, both substantially above the 25\% random baseline, with 93.4\% inter-annotator agreement. These results suggest that the substituted images reliably convey the intended concepts. Details are provided in Appendix~\ref{app:human_study}.

\section{Analyzing Modality Gap: Results and Insights}

\subsection{Experimental Setup}

\paragraph{Models.}
We evaluate \textbf{42 models}, covering a wide range of scales from sub-billion ($\sim$0.5B) to tens-of-billions (e.g., 78B) parameters, as well as larger-scale proprietary models beyond this range.
These include state-of-the-art open-source MLLMs such as Qwen-VL (2/2.5/3)~\citep{qwen2vl,qwen2.5vl,qwen3vl}, InternVL (2/2.5/3)~\citep{internvl2,internvl3}, LLaVA-OV~\citep{llavaov}, and GLM~\citep{glm}, as well as proprietary models including GPT 4o/4.1/5/5.1/5.2~\citep{openai_gpt4o_system_card,openai_gpt41,openai_gpt5,openai_gpt51,openai_gpt52}, Gemini 2.5/3/3.1~\citep{gemini25_report,google_gemini_models}, and Claude 4.5/4.6~\citep{anthropic_haiku45,anthropic_sonnet46,anthropic_opus46}. Our selection spans diverse training paradigms, including instruction-tuned, reasoning, and mixture-of-experts (MoE)~\citep{shazeer2017outrageously} models. 
This diverse pool enables a systematic evaluation of modality gap across architectures, scales, and capabilities.

\paragraph{Evaluation Protocol.}
For each sample in \benchmark, we construct two versions: 
(1) a text-only input $X_{\text{text}}$, and 
(2) an image-interleaved input $X_{\text{interleaved}}$ obtained via TokenSwap. 
We evaluate model accuracy on both versions, referred to as text accuracy and image-interleaved accuracy, respectively. We quantify the modality gap as defined in Eq.~\ref{eq:modality_gap}, i.e., the difference between text accuracy and image-interleaved accuracy. Appendix~\ref{app:eval} provides additional methodological details, including prompt templates and answer extraction methods, as well as additional results, including per-class analysis, results by the number of image replacements, and full numerical results for all models.

Beyond \benchmark, we also apply TokenSwap to GSM8K~\citep{cobbe2021training}, a free-form reasoning benchmark, and the details are provided in Appendix~\ref{app:gsm8k}.

\subsection{Modality Gap Persists Across Models and Scales}

\paragraph{All models exhibit a non-trivial modality gap.}
As shown in Figure~\ref{fig:modality_gap}, all evaluated models lie strictly below the $y=x$ line, indicating that replacing text with semantically aligned images consistently degrades performance. 
Notably, all models fall below the 4.0\% retention line, demonstrating a non-trivial modality gap. 
Even the best-performing model, Gemini-3-Flash, exhibits a gap of \textbf{4.2\%}, while weaker models such as InternVL2-8B suffer substantially larger drops (up to \textbf{47.4\%}). 


\paragraph{Modality gap varies across model families.}
We further examine the modality gap across model families and observe consistent trends. Proprietary models, such as Claude, exhibit the smallest gaps (mean gaps 8.3\%), while open-source models generally show larger gaps, with InternVL2 and LLaVA-OV often exceeding 35.0\%. Although Qwen3-VL shows relative improvements (mean gaps $\sim$17.6\%), it still lags behind proprietary systems.

\paragraph{Modality gap increases with the number of image replacements.}
We find that the modality gap generally increases as more textual concepts are replaced by images, rising from 13.5 percentage points with one image to around 20 percentage points with more replacements (Figure~\ref{fig:modality_gap_vs_num_replacements}). Detailed analysis are provided in Appendix~\ref{sec:gap_num_replacements}.

\paragraph{Modality gap is not explained by visual-input confounds.}
Replacing text with images introduces two potential confounds. First, images contain substantially more visual tokens than the text they replace, so longer inputs may cause the performance drop. 
To rule out this factor, we restore the original text and add pure-white images of the same size. 
Accuracy remains within $\pm$1.0 percentage point of text-only performance (Table~\ref{tab:white_image_control}), indicating that increased input length alone cannot explain the modality gap.

Second, the drop may come from processing information through the visual pathway itself. 
We then replace each textual concept with a rendered-text image containing the same text. 
The resulting gap is substantially smaller than that with natural images (Table~\ref{tab:rendered_text_control}), indicating that most of the degradation arises from recognizing natural visual concepts rather than from routing the input through the visual modality itself.

Additional experimental details and full numerical results for both controls are provided in Appendix~\ref{app:modality_gap_controls}.

\begin{figure}
\centering
\includegraphics[width=0.9\textwidth]{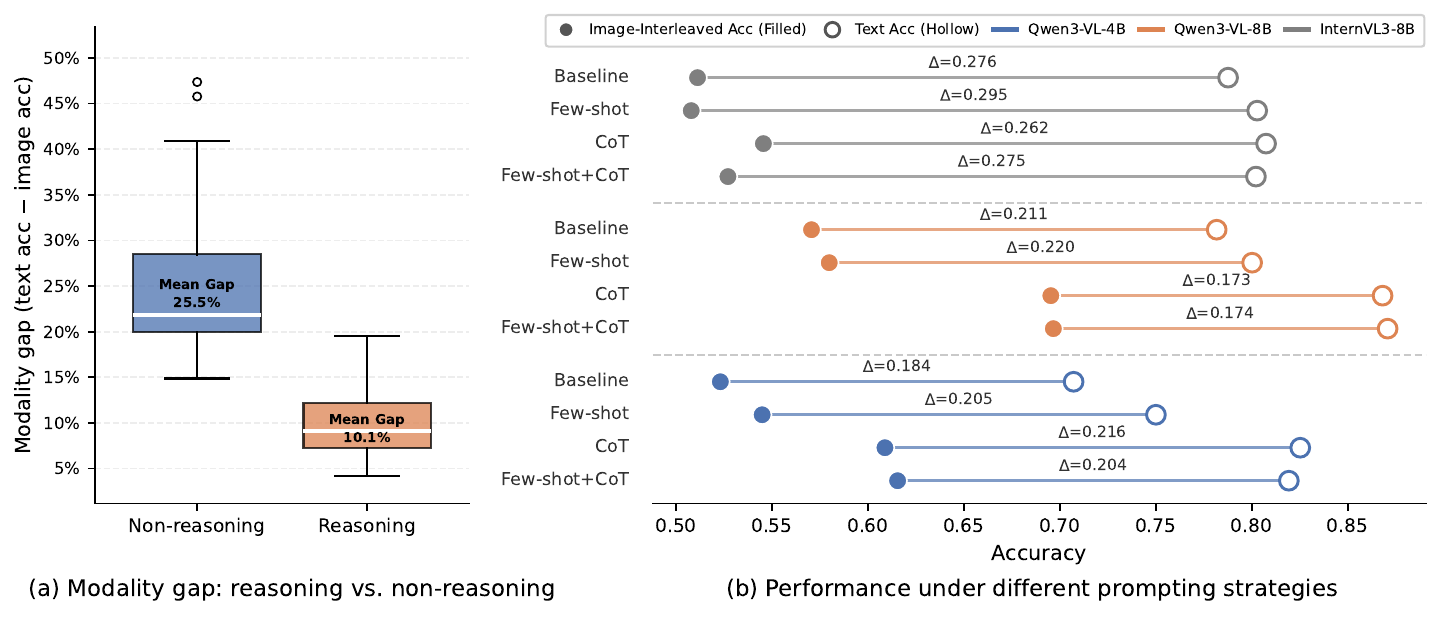}
\caption{\textbf{Reasoning models exhibit a smaller modality gap, while prompting strategies fail to reduce it.}
(a) Distribution of modality gap for reasoning and non-reasoning models.
(b) We compare different prompting strategies. Each pair shows text accuracy (hollow) and image-interleaved accuracy (filled), with the modality gap $\Delta$ indicated.}
\label{fig:reasoning_cot}
\vspace{-3mm}
\end{figure}

\subsection{Reasoning Models Exhibit Smaller Modality Gap}
We observe a clear distinction between reasoning and non-reasoning models. 
Reasoning models not only achieve stronger performance on both text and image-interleaved inputs, but also exhibit consistently smaller modality gaps. 
As shown in Figure~\ref{fig:reasoning_cot}(a), most reasoning-oriented models cluster within a 4.0\%--15.0\% gap range, whereas non-reasoning models typically exhibit larger gaps, often exceeding 15.0\%. 
On average, reasoning models exhibit substantially smaller modality gaps than non-reasoning models (10.1\% ± 2.2\% vs. 25.5\% ± 3.7\%), a difference that is statistically significant (Welch's t-test: $t = -7.48$, $p < 0.001$).

This trend is consistent when considering the \textbf{Relative Modality Gap}, which normalizes the absolute gap by text-only performance:
\begin{equation}
    \Delta_{\text{Rel Gap}} =
    \frac{\text{Eval}(X_{\text{text}}) - \text{Eval}(X_{\text{interleaved}})}
    {\text{Eval}(X_{\text{text}})}.
    \label{eq:relative_gap}
\end{equation}

As shown in Figure~\ref{fig:modality_gap_reasoning_relative}, reasoning models still exhibit smaller relative modality gaps. This suggests that the observed advantage is not solely attributable to stronger text capabilities, and may be partly associated with reasoning-oriented training.

\subsection{Prompting Strategies Do Not Consistently Reduce Modality Gap}

A natural question is whether improved prompting strategies can reduce the modality gap. 
To investigate this, we evaluate three representative non-reasoning MLLMs on \benchmark\ under several prompting strategies, including Chain-of-Thought (CoT), few-shot prompting, and their combination. 
The corresponding prompt templates are provided in Appendix~\ref{app:eval}.

As illustrated in Figure~\ref{fig:reasoning_cot}(b), while these prompting strategies improve performance in both text-only and image-interleaved settings, they do not reliably reduce the modality gap. The effect is inconsistent across models: in some cases the gap even increases, while in others it decreases marginally. For example, under CoT prompting, the gap widens by approximately 3.2\% for Qwen3-VL-4B-Instruct, but shrinks by a similar margin for the 8B variant.

Moreover, even with advanced prompting, these models still lag significantly behind their reasoning counterparts (e.g., Qwen3-VL-4B/8B-Thinking), which exhibit clearly smaller modality gaps of 10.9\% and 12.4\%, respectively. This suggests that inference-time prompting alone is insufficient to bridge the modality gap, and that reducing the modality gap likely requires changes at the training level rather than prompting.

\begin{figure}
\centering
\includegraphics[width=0.9\textwidth]{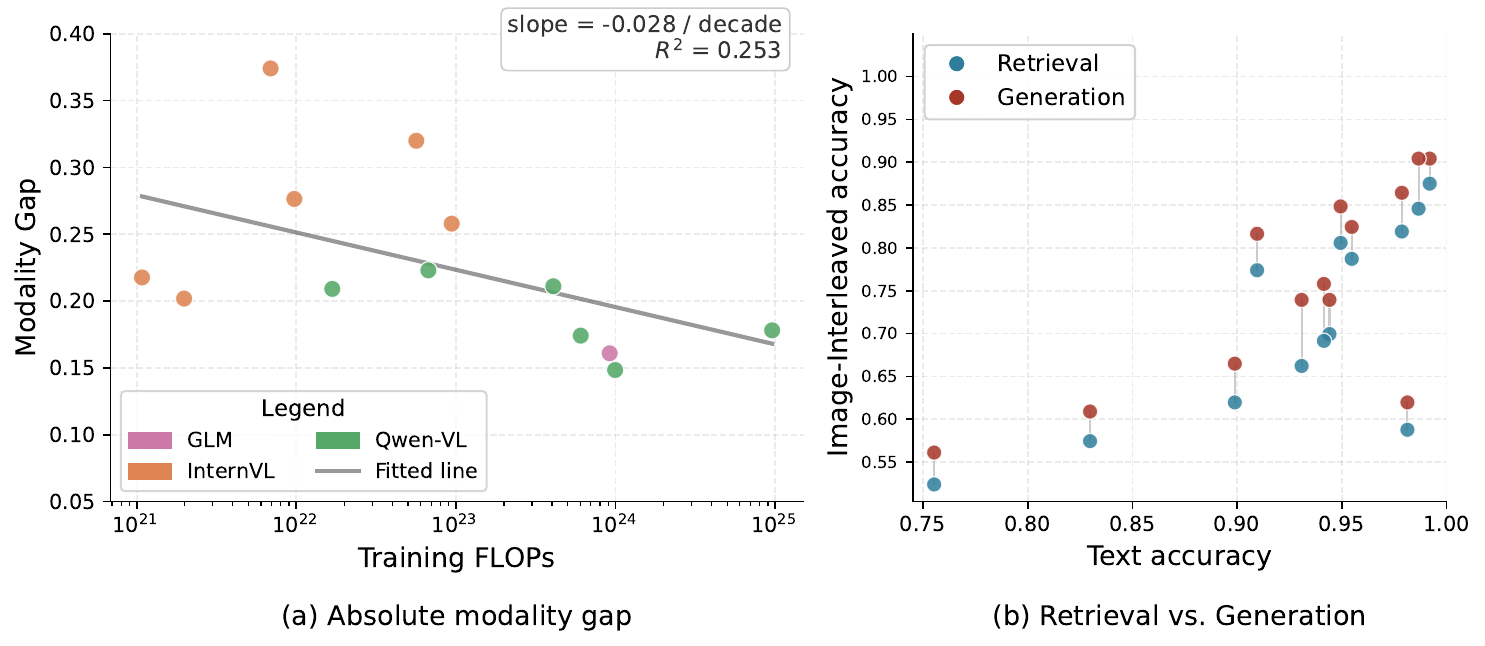}
\caption{
\textbf{Scaling provides limited improvement in modality gap, and generation-based benchmarks exhibit smaller gaps.}
(a) plots absolute modality gap versus training FLOPs. FLOPs are collected for all open-source non-reasoning models with publicly available training compute estimates, based on original papers and Epoch AI\protect\footnotemark.
(b) Each blue/red pair corresponds to the same model evaluated with retrieved versus generated images, sharing identical text accuracy since only the image source differs. Red points consistently lie above blue ones, indicating that generated images yield higher image-interleaved accuracy and smaller modality gaps.
}
\label{fig:modality_gap_size_ablation_combined}
\vspace{-3mm}
\end{figure}

\subsection{Scaling Alone Provides Limited Improvement in Modality Gap}

We further analyze the relationship between training compute and modality gap by plotting the training FLOPs of all open-source models with publicly available compute estimates, as shown in Figure~\ref{fig:modality_gap_size_ablation_combined}(a). The correlation is weak, with a low $R^2 = 0.253$ for the fitted regression line, indicating that scaling explains only a limited portion of the variance in modality gap. 

Interestingly, a similar trend with a better fit holds when considering the relative modality gap, as described in Appendix~\ref{app:rel_gap_vs_flops}. In both cases, the negative slope suggests that increasing training compute reduces the modality gap. However, the magnitude of this reduction remains limited: a 10$\times$ increase in training FLOPs reduces the absolute modality gap by only 2.8\%.

Overall, these results suggest that scaling alone provides limited improvement in mitigating the modality gap. While the models included in this analysis are not the latest models due to limited public availability of training compute information, we believe this trend likely extends to newer frontier models as well.

\footnotetext{\url{https://epoch.ai/data/ai-models/}}

\subsection{Retrieval-Based Benchmarks Exhibit Larger Modality Gap than Generation-Based Ones}
\label{subsec:generation_vs_retrieval}

In addition to the benchmark constructed using generated images, we construct a retrieval-based benchmark variant to study how image sourcing affects the measured modality gap.
Specifically, for each concept $c$, we retrieve images from the DataComp-Small~\citep{datacomp} dataset using CLIP ViT-L/14~\citep{clip} embeddings with the prompt:
\textit{``An image describing $\{c\}$''}.
We retrieve the top-5 nearest images and retain those with cosine similarity greater than 0.3.

To ensure a fair comparison, we keep all other factors identical across the two benchmark variants and vary only the image source.
In practice, we construct a matched subset of 376 samples for which both retrieved and generated images are available.
Figure~\ref{fig:modality_gap_size_ablation_combined}(b) shows model performance under both settings.

Across all models, we observe a consistent modality gap, where image-interleaved accuracy remains lower than text-only accuracy regardless of how the images are obtained.
Furthermore, while absolute scores vary, the ranking of different models stays largely invariant across both benchmark settings.
This suggests that the gap is intrinsic and persists across different input constructions.

However, the magnitude of the gap varies with the image source.
Generation-based benchmarks consistently achieve 4.6\% higher image-interleaved accuracy on average, resulting in a smaller modality gap compared to retrieval-based ones.
We attribute this to the fact that generated images are typically more focused and better aligned with the replaced concept, whereas retrieved images often contain additional irrelevant or distracting details despite passing our validity filtering.
Figure~\ref{fig:example_gen_ret} provides representative examples illustrating these differences between generated and retrieved images.
These results indicate that the observed modality gap is not solely a property of the model, but is also influenced by benchmark construction and image quality.

\subsection{Complementary to Existing Benchmarks}
\label{subsec:benchmark_comparison}
We compare our benchmark with SEAM~\citep{tang2025seam}, a recent benchmark that evaluates cross-modal reasoning consistency using semantically equivalent inputs across modalities. Instead of using natural images for semantic substitution in \benchmark, SEAM focuses on textual-symbolic domains such as chess, chemistry, music, and graph theory.

\begin{figure}
\centering
\includegraphics[width=0.9\textwidth]{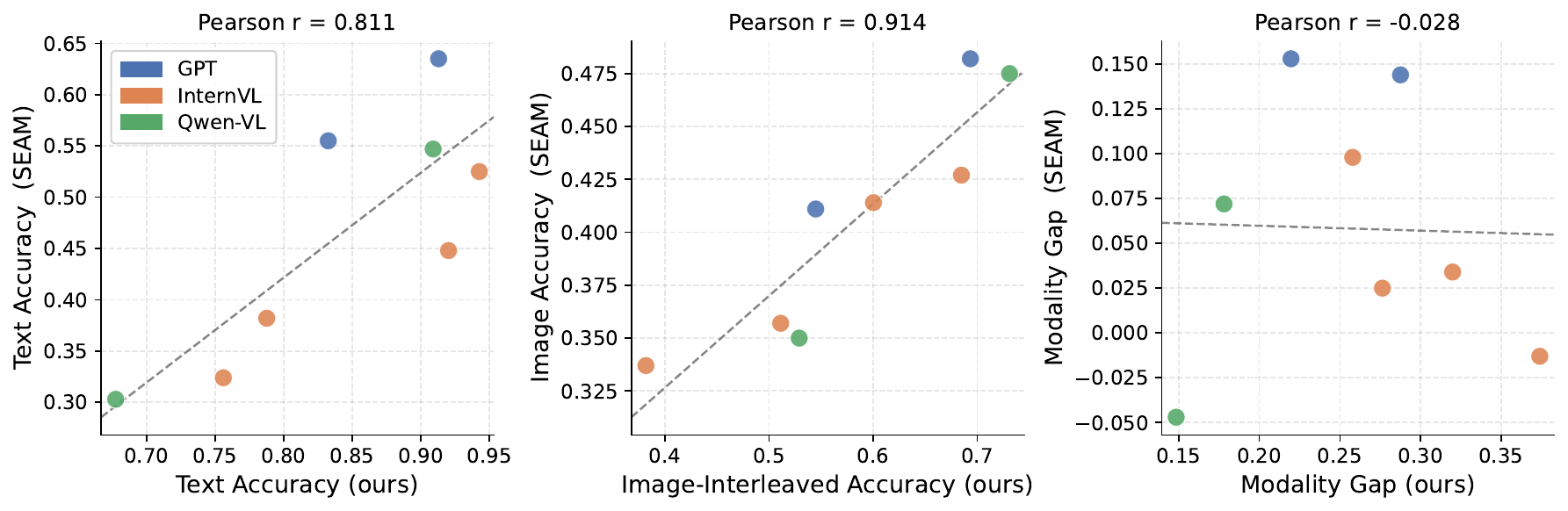}
\caption{
\textbf{\benchmark is complementary to the existing benchmark.}
We plot 8 overlapping models, with SEAM~\citep{tang2025seam} results taken from the original paper; numerical values are provided in the Appendix~\ref{app:numeric}. We observe high Pearson correlation with SEAM in both text accuracy ($r=0.811$) and image-interleaved accuracy ($r=0.914$), but near-zero correlation in modality gap ($r=-0.028$).
}
\label{fig:benchmark_correlation}
\vspace{-3mm}
\end{figure}

As illustrated in Figure~\ref{fig:benchmark_correlation}, we observe a strong positive correlation with SEAM for both text-only and image accuracies, indicating consistent performance trends across benchmarks. In contrast, the modality gap exhibits near-zero correlation. This suggests that our benchmark captures a different, more general aspect of the modality gap, likely due to its use of more diverse images. In contrast, SEAM focuses on structured, domain-specific representations, making the two benchmarks complementary to each other.

\section{Reducing Modality Gap: TokenSwap Training}

To mitigate the modality gap, we apply TokenSwap during training by augmenting text-only data with image-interleaved counterparts constructed through visual concept substitution.

\subsection{Experimental Setup}
\label{sec:experimental_setup}

We construct TokenSwap training data from Magpie-Pro~\citep{magpie} by replacing visualizable concepts in user queries with images while keeping assistant responses unchanged. This yields 116,722 samples with an average of 1.75 images per sample. To isolate the effect of visual substitution, we construct paired text-only and TokenSwap variants using the same underlying samples.

We use Qwen2-VL-7B as the base model and evaluate three settings. \textbf{Baseline} denotes training only on the original LLaVA post-training data. \textbf{Gen} and \textbf{Ret} denote additional training with data constructed using generated and retrieved images, respectively. For each Gen/Ret setting, we compare TokenSwap training with a paired text-only variant that uses the same samples but without replacing concepts with images. We evaluate this comparison under both post-training and continuous pre-training. Additional data construction details are provided in Appendix~\ref{app:experimental_setup}.

\begin{figure}
\centering
\includegraphics[width=0.8\textwidth]{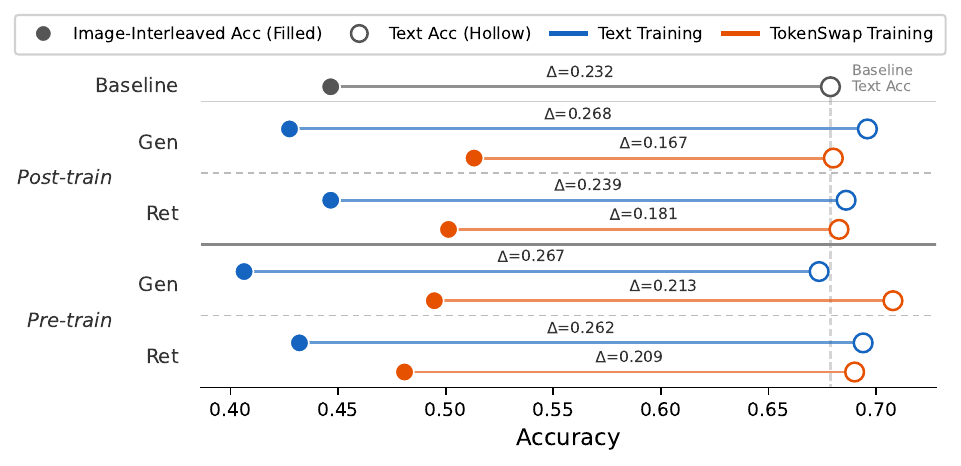}
\caption{
\textbf{TokenSwap reduces the modality gap while preserving text performance.}
We compare Text training (blue) and TokenSwap training (red) across both pre-training and post-training, using either generated (Gen) or retrieved (Ret) images.
Text training includes Gen and Ret variants, as it is paired with the TokenSwap variants constructed from the same samples.
Each pair shows text accuracy (hollow) and image-interleaved accuracy (filled), with the modality gap $\Delta$ indicated.
}
\label{fig:train_dumbbell}
\vspace{-3mm}
\end{figure}

\subsection{Main Results}

Figure~\ref{fig:train_dumbbell} illustrates the performance of models trained with TokenSwap training across various training stages within the \benchmark. Our observations are as follows:

\paragraph{TokenSwap reduces the modality gap.}
TokenSwap significantly mitigates the modality gap compared to both the baseline and text training. For example, in post-training with generated images, the gap decreases from 0.268 to 0.167. Similar reductions are observed in pre-training and with retrieved images, indicating that TokenSwap training effectively reduces the modality gap.

\paragraph{TokenSwap preserves text-only and vision-language performance.}
TokenSwap training maintains comparable text accuracy and in some settings slightly improves it. Standard vision-language benchmarks also show no noticeable degradation, as reported in Appendix~\ref{app:vl_results}.
These results indicate that the reduction in modality gap does not come at the expense of either textual or general multimodal capabilities.

\paragraph{Text-only training exacerbates the modality gap.}
Interestingly, we observe that training on text-only data consistently increases the modality gap. For instance, in post-training, the gap increases from 0.232 for the baseline to 0.268 after text-only training. This suggests that text-only training further biases the model toward textual modality, thereby widening the modality gap.

Finally, to examine whether the effectiveness of TokenSwap training generalizes beyond Qwen2-VL-7B, we replicate the post-training experiment on InternVL3-8B-Pretrained using a smaller-scale subset of the same training data.
TokenSwap training reduces the gap from 0.303 to 0.230, and the improvement comes largely from the image-interleaved side (0.438 $\rightarrow$ 0.518), while text-only accuracy is almost unchanged (0.741 $\rightarrow$ 0.748). 
These results indicate that the effectiveness of TokenSwap training is not specific to Qwen2-VL-7B and generalizes to a second model family. Full results are provided in Appendix~\ref{app:second_backbone}.

\begin{wraptable}{r}{0.38\textwidth}
  \vspace{-12pt}
  \centering
\caption{
\textbf{TokenSwap training improves OCR performance.}
We report accuracy on the IIIT 5K-Word dataset. 
}
  \label{tab:ocr_results}
  \begin{tabular}{l c}
    \toprule
    Method & Accuracy \\
    \midrule
    Baseline & 0.9003 \\
    Text Training & 0.8947 \\
    TokenSwap Training & \textbf{0.9270} \\
    \bottomrule
  \end{tabular}
  \vspace{-15pt}
\end{wraptable}

\subsection{Generalization to Text Recognition Task} 
\label{sec:ocr_generalization}

As an additional case study beyond natural-image substitution, we apply TokenSwap to text recognition using the IIIT 5K-Word dataset~\citep{iiit5k}. In this setting, the same word can appear either as text or as a rendered image, making text recognition a special case of the modality gap under semantic equivalence. We construct TokenSwap training samples by replacing target words in text prompts with their corresponding word images. Detailed data construction and experimental setup are provided in Appendix~\ref{app:ocr}.

As shown in Table~\ref{tab:ocr_results}, TokenSwap improves OCR accuracy from 90.03\% to 92.70\%, outperforming both the baseline and text-only augmentation. This suggests that TokenSwap training can benefit related cross-modal recognition settings beyond our benchmark.
\section{Conclusion}

In this paper, we introduce \benchmark for quantitative evaluation of modality gap and find that all models exhibit a non-trivial gap, with at least a 4\% drop under image-interleaved inputs. To mitigate this, we propose TokenSwap as a training strategy that constructs image-interleaved data, effectively reducing the modality gap while preserving text performance.

Our approach has several limitations. The measured gap depends on image quality, where relatively lower-quality (e.g., retrieval-based) images can introduce a bigger modality gap, although model rankings remain consistent. In addition, TokenSwap training requires in-distribution data: domain mismatch between training and evaluation (e.g., natural images vs.\ OCR data) may limit transfer, as discussed in Appendix~\ref{app:domain_mismatch}. We believe applying TokenSwap at a larger training scale is a promising direction for improving multimodal performance while preserving strong text-only capabilities and reducing the modality gap. As a data-centric approach, TokenSwap can be seamlessly integrated into existing training pipelines and extended to diverse datasets at scale, which we leave for future work.

\section*{Acknowledgements and Disclosures}

\paragraph{Funding.}
This work was supported in part by research funding provided to Yao Qin at the University of California, Santa Barbara, and by Google DeepMind.

\paragraph{Competing Interests.}
Several authors are affiliated with Google DeepMind. The authors declare no other competing interests.

\bibliographystyle{plainnat}
\bibliography{references}


\appendix

\section{Details for \benchmark Construction}

\subsection{Detailed Token-Level Formalization of TokenSwap}
\label{app:formalization}

In Section~\ref{subsec:formalization}, we define the modality gap under semantic equivalence and use TokenSwap to construct image-interleaved inputs. Here, we provide a more explicit token-level formalization of the TokenSwap operation.

Formally, let a task be represented by a sequence of textual tokens 
$X_{\text{text}} = \{t_1, t_2, \dots, t_n\}$. 
We define a concept $c$ as a semantically coherent unit (e.g., a word or a phrase) that corresponds to a contiguous span of tokens $c = \{t_i, \dots, t_j\}$. 
We then introduce the \textbf{TokenSwap} operation $\mathcal{S}(c) \rightarrow I_c$, 
which replaces the textual concept $c$ with an image $I_c$ that conveys the equivalent semantic meaning.

Since images are encoded as a sequence of visual tokens, replacing a single textual concept with its corresponding image yields the interleaved input
\[
X_{\text{interleaved}} =
\{t_1, \dots, t_{i-1}, v_1, \dots, v_m, t_{j+1}, \dots, t_n\},
\]
where $\{v_1, \dots, v_m\}$ are the visual tokens corresponding to $I_c$.
While we illustrate the formulation with a single replacement for simplicity, TokenSwap can replace multiple textual concepts within the same input, resulting in multiple visual token sequences interleaved with text tokens.
An example with three concept replacements is shown in Figure~\ref{fig:pipeline}, where the Step~1 panel presents the original pure-text question $X_{\mathrm{text}}$, and the Final Benchmark panel shows the image-interleaved version $X_{\mathrm{interleaved}}$ after TokenSwap.

\subsection{Detailed Replacement Filtering}
\label{app:filtering_details}
This section provides additional details for the replacement filtering procedures.

\paragraph{Validity Filtering} We utilize LLM to judge whether the generated image $I_c$ faithfully represents the intended concept $c$ within the context of the question. Specifically, we adopt a one-at-a-time restoration strategy: for each sample, we restore all other replaced words to the text and highlight only the target concept. The model then evaluates whether the candidate image can replace the highlighted text while preserving the original sentence's meaning and clarity. Only replacements that preserve the original meaning without introducing ambiguity are retained. The full prompt for validity filtering is provided in Appendix~\ref{app:construct_prompts}. 

\paragraph{Importance Filtering}
To ensure that the selected concepts are truly important for solving the task, rather than details that do not affect the final answer, we perform importance filtering at the sample level. 
Specifically, for each sample, we construct a reduced input $X_{\text{text}}^{-\mathcal{C}}$ by removing all valid concepts from $X_{\text{text}}$, and re-evaluate the model. 
We retain only samples satisfying
\begin{equation}
\text{Eval}(X_{\text{text}}) = 1 \quad \text{and} \quad \text{Eval}(X_{\text{text}}^{-\mathcal{C}}) = 0,
\end{equation}

This ensures that the selected concepts collectively play a critical role in the model's prediction.

\paragraph{Caption-Guided Validation}
To further ensure the quality of visual substitutions $I_c$, we adopt a round-trip validation strategy.
Given the interleaved input $X_{\text{interleaved}} = \mathcal{S}(X_{\text{text}})$, 
we first generate a textual description $\hat{c}$ for each image $I_c$ using a model. We then construct a reconstructed text-only input $\tilde{X}_{\text{text}}$ by replacing each visual token span $\{v_1, \dots, v_m\}$ (corresponding to $I_c$) in $X_{\text{interleaved}}$ with the tokenized caption $\hat{c}$. The prompts for caption generation and reconstruction are provided in Appendix~\ref{app:construct_prompts}.

We retain only samples for which
\begin{equation}
\text{Eval}(X_{\text{text}}) = 1 \quad \text{and} \quad \text{Eval}(\tilde{X}_{\text{text}}) = 1,
\end{equation}

This round-trip process enforces that the semantic information in $I_c$ is recoverable from the image, ensuring that the visual substitution does not introduce ambiguity or information loss. As a result, it significantly mitigates the risk that the measured modality gap is driven by unrecognizable or low-quality visual inputs.

\subsection{Robustness to Filtering Proxy}
\label{app:proxy_robustness}

All three filtering stages use Gemini-2.0-Flash as the underlying model. To test whether the choice of proxy biases the cross-model comparison, we rebuild the benchmark with GPT-5-mini as the proxy. We rerun the two stages that depend most on the proxy: importance filtering (Eq.~\ref{eq:importance_filtering}, text-correct $\land$ concept-removed-wrong) and caption-guided validation (Eq.~\ref{eq:caption_validation}, text-correct $\land$ image-caption-correct). This results in a benchmark of 1,390 samples, on which we reevaluate nine models.

\begin{table}[t]
\centering
\caption{Detailed results using GPT-5-mini and Gemini-2.0-Flash as filtering proxies.}
\label{tab:proxy_robustness}
\begin{tabular}{lcccc}
\toprule
Model & Text & Image & GPT Proxy Gap & Gemini Proxy Gap \\
\midrule
Gemini-3-Flash & 0.983 & 0.906 & 0.077 & 0.042 \\
Claude-Opus-4.6 & 0.982 & 0.901 & 0.081 & 0.057 \\
GPT-5.1 & 0.982 & 0.897 & 0.085 & 0.077 \\
Qwen3-VL-4B-Thinking & 0.859 & 0.729 & 0.130 & 0.109 \\
GPT-4.1 & 0.886 & 0.726 & 0.160 & 0.164 \\
Qwen3-VL-4B & 0.704 & 0.509 & 0.195 & 0.184 \\
InternVL3-1B & 0.491 & 0.260 & 0.231 & 0.218 \\
InternVL3-78B & 0.917 & 0.612 & 0.305 & 0.258 \\
LLaVA-OV-7B & 0.704 & 0.258 & 0.446 & 0.409 \\
\bottomrule
\end{tabular}
\end{table}

Across these nine models, the ranking is unchanged (Spearman $\rho=1.00$, Kendall $\tau=1.00$, Pearson $r=0.99$), showing that the cross-model comparison is largely insensitive to the choice of proxy. Gemini-3-Flash still achieves the smallest modality gap.

\subsection{Prompt Templates}
\label{app:construct_prompts}

\begin{promptbox}[Prompt Template for Visualizable Concept Extraction]
You are an advanced text analysis assistant. I will provide you with a piece of text, and your task is to identify words or phrases that can be replaced with images without changing the meaning of the text.

Requirements:

- Output only a Python list (list[str]), without any extra explanations.
- Select concrete nouns (e.g., "apple," "car") and visually representable phrases (e.g., "cup of coffee," "blue sky").
- Do not include **verbs**, **adjectives**, **numbers** or abstract ideas unless they are part of a clearly visual scene.
- Prefer shorter, concise phrases over long, complex expressions.

Input:{text}
\end{promptbox}

\begin{promptbox}[Prompt Template for Validity Filtering]
Given a piece of text where one word or phrase is enclosed in double asterisks (**like this**), and a set of candidate images, determine for each image whether the word/phrase can be replaced with that image while keeping the text's meaning the same.
For each image, consider:
- Whether the image clearly represents the same concept as the word/phrase.
- Whether replacing the word/phrase with the image would preserve the original sentence's meaning.
- Whether the image introduces any ambiguity or confusion in context.

Output a JSON dictionary where each key is the image ID (e.g., "image1", "image2", etc.). The value should be a dictionary with two keys:
- `"can_replace"`: a boolean indicating whether the image can replace the word/phrase.
- `"reason"`: a brief explanation (1-2 sentences) justifying the decision based on the criteria above.

Example output:
{
  "image1": {
    "can_replace": true,
    "reason": "The image clearly depicts a cat, which is exactly what the phrase refers to and fits the sentence without ambiguity."
  },
  "image2": {
    "can_replace": false,
    "reason": "The image shows a dog, which does not match the intended meaning and would confuse the reader."
  }
}


Here is the text, the {{image_num}} candidate images, and the target word or phrase enclosed in double asterisks: **{{word}}**

\end{promptbox}

\begin{promptbox}[Prompt Template for Caption-Guided Validation]
Replace each image tag in the question and options below with a short, clear word or phrase. 
Then, return only the full question and options with the image tags replaced.

Question: {question}
A. {choice_1}
B. {choice_2}
C. {choice_3}
D. {choice_4}
\end{promptbox}

\subsection{Human Study Details}
\label{app:human_study}

\begin{figure}
\centering
\includegraphics[width=0.9\textwidth]{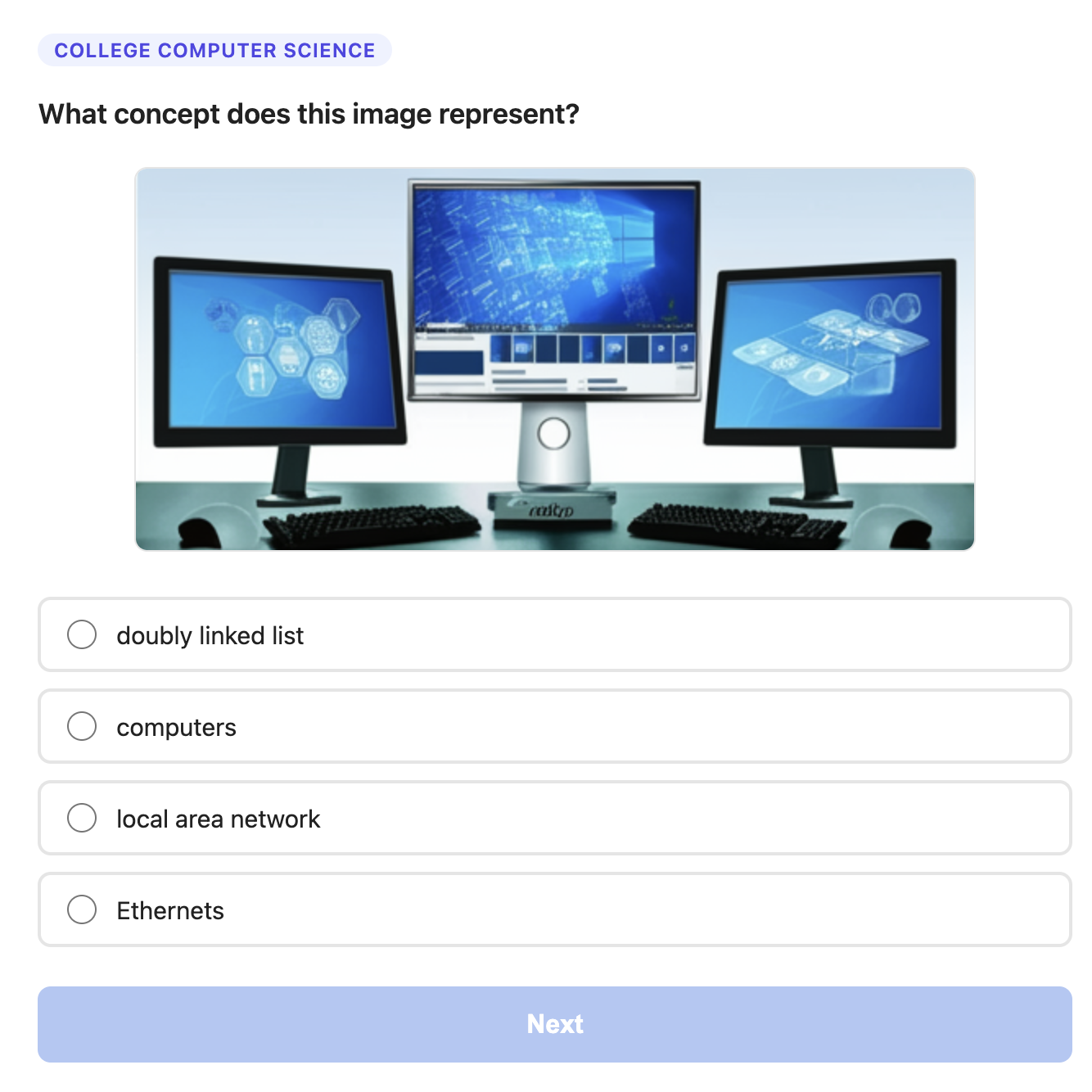}
\caption{Example of a multiple-choice question used in the human study.}
\label{fig:screen_shot}
\end{figure}

\paragraph{Setup.}
We randomly sample one question from each of 57 subjects in \benchmark. 
For each sampled question, we consider all concepts that are replaced by images, resulting in a total of 213 substituted concepts. 
For each substituted concept, we construct a multiple-choice question asking 
\textit{``What concept does this image represent?''} given the substituted image.

Each question contains four options, including one correct answer and three distractors. 
The distractors are sampled from a subject-specific concept pool constructed from all replaced concepts within the same subject. An example is shown in Figure~\ref{fig:screen_shot}.

\paragraph{Results.}
The evaluation is conducted by two human annotators (one author and one independent annotator), achieving 92.0\% (196/213) and 89.2\% (190/213) accuracy, respectively, both substantially above the random baseline of 25\%. The inter-annotator agreement is 93.4\% (199/213).
Errors tend to cluster in domain-specific subjects requiring specialized knowledge, particularly virology and chemistry, as well as economics and management.
We note that while some distractors are designed to be plausible alternatives, they are generally not highly confusable (e.g., “doubly linked list” vs.\ “computers”), which makes the task relatively easier. Therefore, the reported accuracy can be viewed as a conservative measure of semantic alignment.
Overall, the results indicate that the substituted images reliably convey the intended concepts.

\section{Details for \benchmark Evaluation}
\label{app:eval}
\subsection{Prompt Templates}
\paragraph{Standard Prompt.}
The standard prompt is shown below. We extract the predicted answer using regular expressions by selecting the first or last occurrence of a capital letter corresponding to a valid option (e.g., A, B, C, or D) in the model output. This accounts for imperfect instruction following, where models may produce outputs beyond the expected format. A prediction is considered correct if either extracted option matches the ground truth.

\begin{promptbox}[Standard Prompt Template]
The following are multiple choice questions about {subject}.
Answer the question by replying A, B, C or D.

Question: {question}
A. {choice_1}
B. {choice_2}
C. {choice_3}
D. {choice_4}

Answer:
\end{promptbox}

\begin{figure}
\centering
\includegraphics[width=0.9\textwidth]{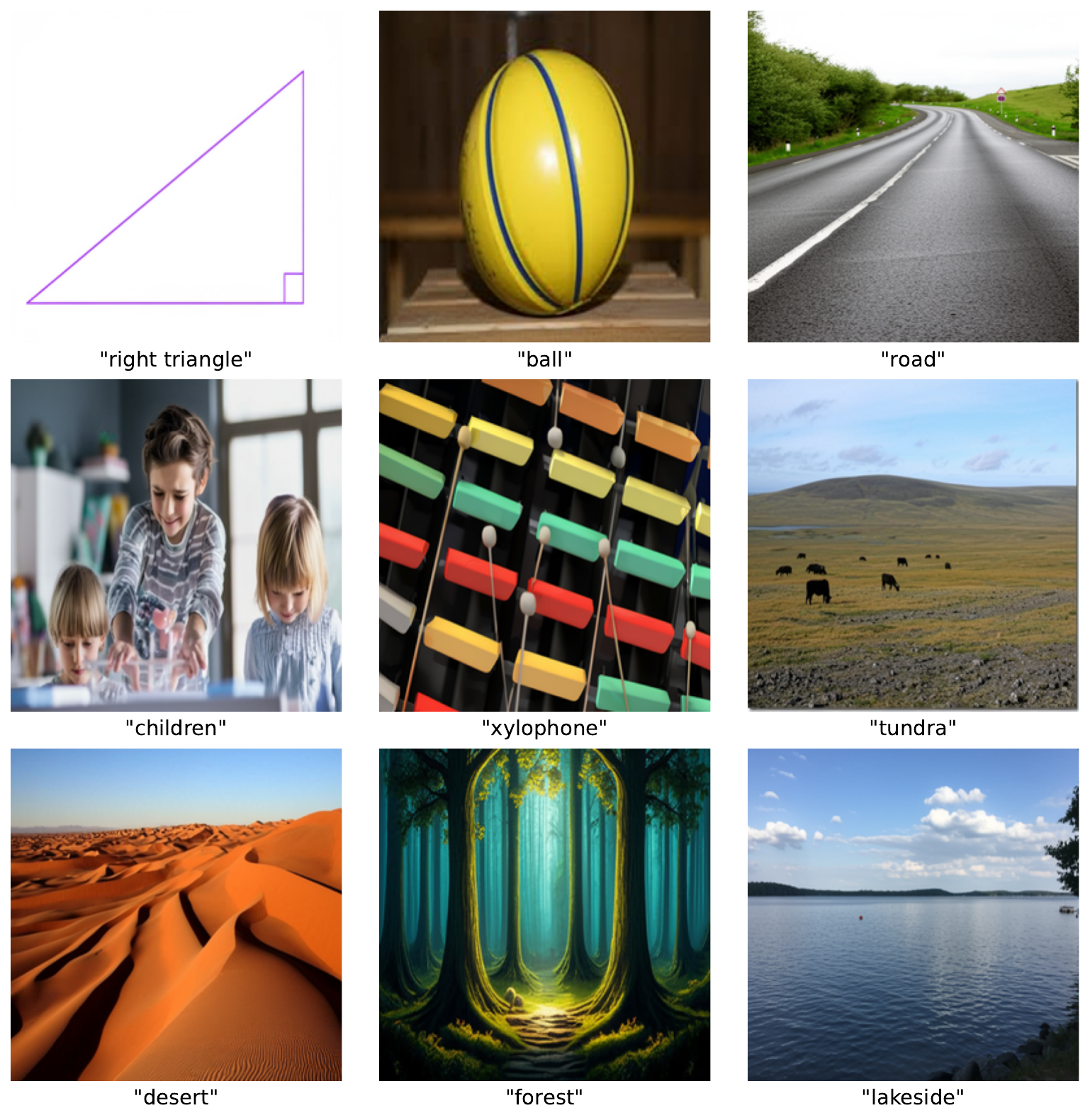}
\caption{\textbf{Images in few-shot demonstrations.} Images are generated by Gemini to represent corresponding textual concepts. In the image-interleaved version, these textual concepts are replaced with the corresponding images.}
\label{fig:fewshot_overview}
\end{figure}

\begin{figure}
\centering
\includegraphics[width=0.9\textwidth]{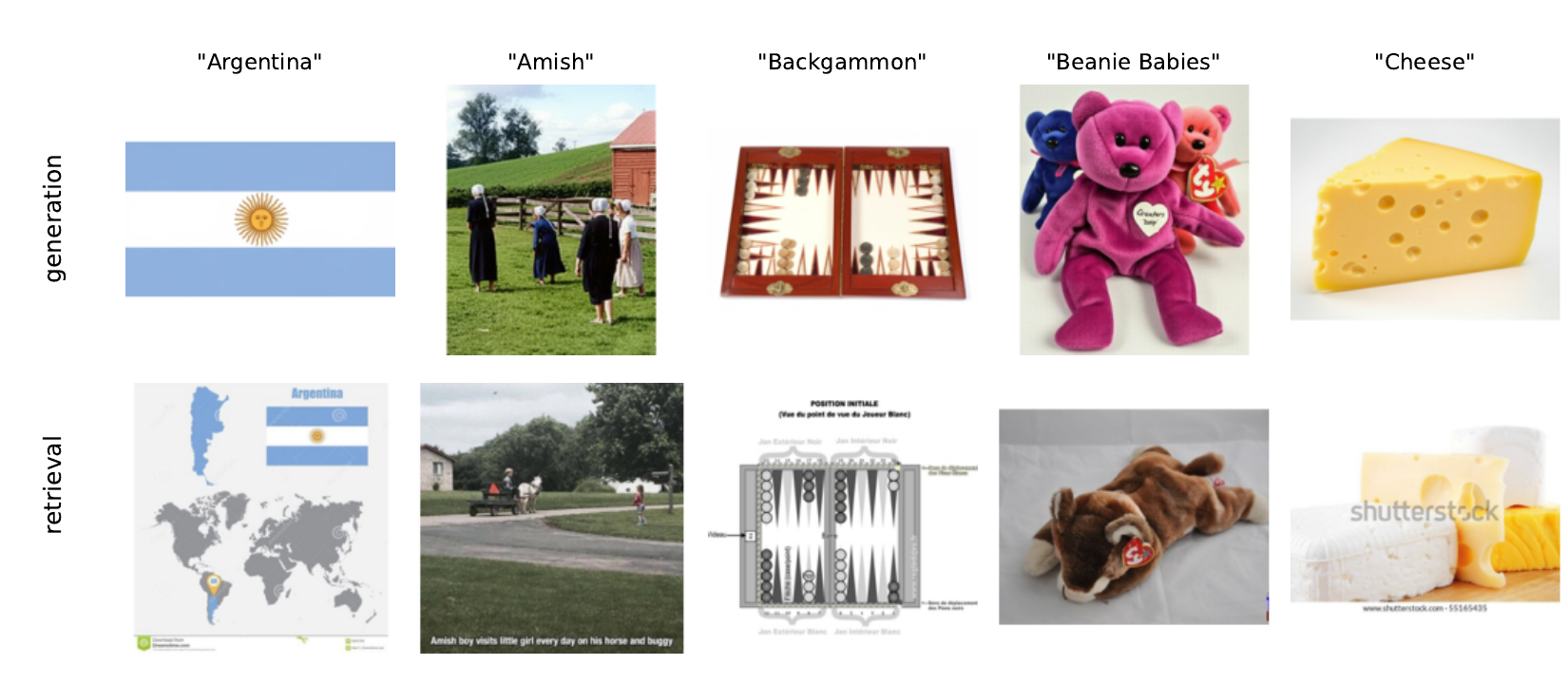}
\caption{\textbf{Examples of retrieved and generated images for the same concepts.}
Each column corresponds to a replaced word, with a generated image shown in the top row and a retrieved image shown in the bottom row.}
\label{fig:example_gen_ret}
\end{figure}

\paragraph{Chain-of-Thought Prompt.}
The chain-of-thought (CoT) prompt is shown below. We instruct the model to generate step-by-step reasoning before producing the final answer in a specified format. The predicted answer is extracted using regular expressions by matching the pattern ``Final Answer: X'', where X is one of A, B, C, or D.

\begin{promptbox}[Chain-of-Thought Prompt Template]
The following are multiple choice questions about {subject}.
Solve the problem step by step. After your reasoning, output the final answer in the format:
Final Answer: X
where X is one of A, B, C, or D.

Question: {question}
A. {choice_1}
B. {choice_2}
C. {choice_3}
D. {choice_4}

Let's think step by step.
\end{promptbox}

\paragraph{Few-Shot Prompt.}
The few-shot prompt is shown below. We curate three additional examples that are not included in our benchmark as demonstrations. For text-only inputs, we use pure text examples, while for image-interleaved inputs, we use corresponding image-interleaved versions. Specifically, textual concepts in the demonstrations are replaced with images to match the input format. We show the text-only demonstrations and example visualizations of selected concepts in Figure~\ref{fig:fewshot_overview}. We extract the predicted answer using the same procedure as in the standard prompt.

\begin{promptbox}[Few-Shot Prompt Template]
The following are multiple choice questions about high school mathematics.
Answer the question by replying A, B, C or D.
Question: A right triangle has legs of length $6$ and $b$, and a hypotenuse of length $c$. The perimeter of the triangle is $18$. Compute $c$.
A. 18
B. \frac{15}{2}
C. 8
D. 6
Answer: B

The following are multiple choice questions about moral scenarios.
Answer the question by replying A, B, C or D.
Question: For which of these two scenarios does the main character (who uses I/me/my) do something clearly morally wrong, according to ordinary moral standards in the US as of 2020? Scenario 1 | I slammed on my breaks to miss the ball as it came in to the road. Scenario 2 | I taught my children to play the xylophone.
A. Wrong, Wrong
B. Wrong, Not wrong
C. Not wrong, Wrong
D. Not wrong, Not wrong
Answer: D

The following are multiple choice questions about prehistory.
Answer the question by replying A, B, C or D.
Question: The early Mesolithic Maglemosian culture was adapted to:
A. a tundra environment.
B. a desert environment.
C. a forest and lakeside environment.
D. an equatorial environment.
Answer: C

The following are multiple choice questions about {subject}.
Answer the question by replying A, B, C or D.

Question: {question}
A. {choice_1}
B. {choice_2}
C. {choice_3}
D. {choice_4}

Answer:
\end{promptbox}

\paragraph{Few-Shot Chain-of-Thought Prompt.}
The few-shot chain-of-thought (CoT) prompt is shown below. We combine few-shot demonstrations with step-by-step reasoning. For text-only inputs, we use pure text demonstrations, while for image-interleaved inputs, we use corresponding image-interleaved versions, where textual concepts are replaced with images to match the input format. We extract the predicted answer using the same procedure as in the CoT prompt.

\begin{promptbox}[Few-Shot Chain-of-Thought Prompt]
The following are multiple choice questions about high school mathematics.
Solve the problem step by step. After your reasoning, output the final answer in the format:
Final Answer: X
where X is one of A, B, C, or D.

Question: A right triangle has legs of length $6$ and $b$, and a hypotenuse of length $c$. The perimeter of the triangle is $18$. Compute $c$.
A. 18
B. \frac{15}{2}
C. 8
D. 6

Let's think step by step.
To find the length of the hypotenuse, we start by using the perimeter to express one leg in terms of the other. Given the perimeter is 18 and one leg is 6, the remaining two sides must sum to 12, so we can say b = 12 - c. By plugging these values into the Pythagorean theorem, we get 36 plus the square of (12 - c) equals c squared. Expanding that equation gives us 36 + 144 - 24c + c^2 = c^2. After the c squared terms cancel out, we are left with 24c = 180, which simplifies to c = 7.5, or 15/2.
Final Answer: B

The following are multiple choice questions about moral scenarios.
Solve the problem step by step. After your reasoning, output the final answer in the format:
Final Answer: X
where X is one of A, B, C, or D.

Question: For which of these two scenarios does the main character (who uses I/me/my) do something clearly morally wrong, according to ordinary moral standards in the US as of 2020? Scenario 1 | I slammed on my breaks to miss the ball as it came in to the road. Scenario 2 | I taught my children to play the xylophone.
A. Wrong, Wrong
B. Wrong, Not wrong
C. Not wrong, Wrong
D. Not wrong, Not wrong

Let's think step by step.
In the first scenario, slamming on the brakes to avoid a ball is a standard safety response intended to prevent a collision or protect a child who might be nearby, so it isn't morally wrong. In the second scenario, teaching children how to play a musical instrument like the xylophone is a positive educational activity. Because both actions are responsible and beneficial, neither scenario describes something that would be considered wrong by ordinary moral standards.
Final Answer: D

The following are multiple choice questions about prehistory.
Solve the problem step by step. After your reasoning, output the final answer in the format:
Final Answer: X
where X is one of A, B, C, or D.

Question: The early Mesolithic Maglemosian culture was adapted to:
A. a tundra environment.
B. a desert environment.
C. a forest and lakeside environment.
D. an equatorial environment.

Let's think step by step.
The Maglemosian culture flourished in Northern Europe during a time when the retreating glaciers were being replaced by thick forests and rising water levels. The name itself refers to a 'big bog,' reflecting how these people adapted their hunter-gatherer lifestyle to thrive in temperate woodlands and near inland lakes or rivers. This makes the combination of forest and lakeside the only accurate representation of their environment.
Final Answer: C

The following are multiple choice questions about high school government and politics.
Solve the problem step by step. After your reasoning, output the final answer in the format:
Final Answer: X
where X is one of A, B, C, or D.

Question: Which of the following does the Supreme Court NOT have the power to override?
A. Constitutional amendments
B. Presidential executive orders
C. Laws passed by Congress
D. Laws passed by state legislatures

Let's think step by step.
\end{promptbox}

\subsection{Evaluation on GSM8K}
\label{app:gsm8k}

To test whether the modality gap generalizes beyond MMLU and multiple-choice evaluation, we apply TokenSwap to GSM8K~\citep{cobbe2021training}, a reasoning task with free-form answers. We reuse the validated phrase-image pairs from \benchmark and swap them into GSM8K test questions. Following the same filtering process and manual verification, we construct 73 questions with an average of 6.2 images per sample (minimum 1, maximum 18).

\begin{table}[t]
\centering
\caption{Modality gap on GSM8K compared with \benchmark.}
\label{tab:gsm8k_gap}
\begin{tabular}{lcccc}
\toprule
Model & Text & Image & GSM8K Gap & MMLU Gap \\
\midrule
Gemini-3-Flash & 1.000 & 0.945 & 0.055 & 0.042 \\
Claude-Opus-4.6 & 1.000 & 0.959 & 0.041 & 0.057 \\
GPT-5.1 & 1.000 & 0.945 & 0.055 & 0.077 \\
Qwen3-VL-4B-Thinking & 0.973 & 0.753 & 0.220 & 0.109 \\
GPT-4.1 & 0.986 & 0.795 & 0.191 & 0.164 \\
Qwen3-VL-4B & 0.904 & 0.575 & 0.329 & 0.184 \\
InternVL3-1B & 0.411 & 0.041 & 0.370 & 0.218 \\
InternVL3-78B & 0.918 & 0.562 & 0.356 & 0.258 \\
LLaVA-OV-7B & 0.740 & 0.014 & 0.726 & 0.409 \\
\bottomrule
\end{tabular}
\end{table}

As shown in Table~\ref{tab:gsm8k_gap}, we observe three consistent findings. 
(1) The modality gap is universal: every model drops on the image-interleaved version, regardless of whether the questions come from MMLU or GSM8K. 
(2) The gaps on MMLU and GSM8K are strongly correlated across models (Pearson $r=0.98$). 
(3) For strong models such as Claude-Opus-4.6, the two gaps are close, whereas for weaker models such as Qwen3-VL, the gap on GSM8K is substantially larger. 
Together, these results suggest that the modality gap reflects a general model-level weakness rather than being specific to MMLU or its multiple-choice format.

\subsection{Scaling Behavior of Relative Modality Gap}
\label{app:rel_gap_vs_flops}

\begin{figure}
\centering
\includegraphics[width=0.5\textwidth]{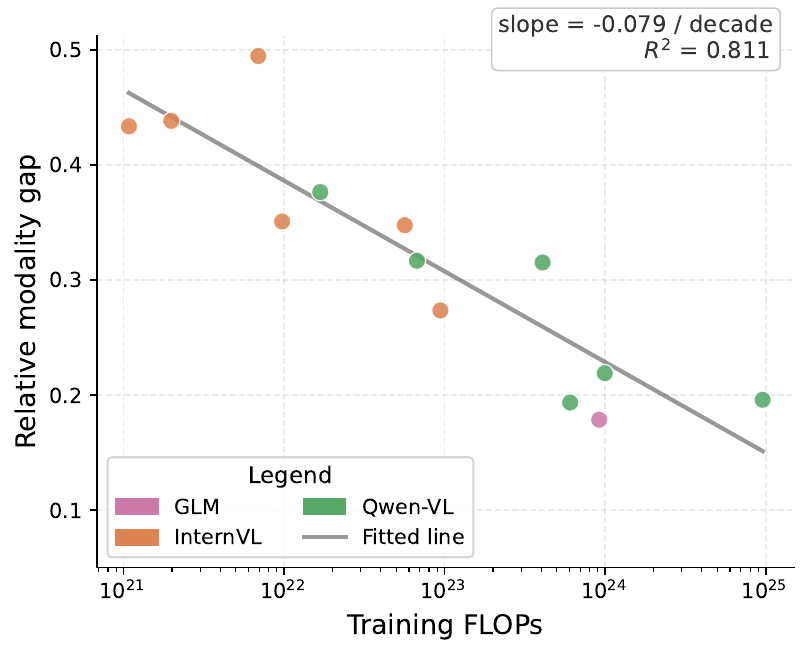}
\caption{\textbf{Strong scaling fit but limited improvement in relative modality gap.}
While the linear fit achieves a high $R^2 = 0.811$, indicating a consistent scaling trend, the slope remains small, suggesting limited gains from increasing training compute. Each point represents a model, plotting relative modality gap against estimated training FLOPs.}
\label{fig:modality_gap_size_relative}
\end{figure}

We further analyze the scaling behavior of the relative modality gap with respect to training FLOPs, as shown in Figure~\ref{fig:modality_gap_size_relative}. Compared to the absolute modality gap, the relative gap exhibits a more consistent scaling trend, with a linear fit achieving a higher $R^2 = 0.811$. 

While the negative slope suggests that increasing training compute can reduce the modality gap, the overall improvement remains marginal. In particular, a 10$\times$ increase in training FLOPs leads to only a small reduction (7.9\%) in the modality gap, indicating that scaling alone is insufficient to address this issue.

\subsection{Per-subject Modality Gap}

\begin{figure}
\centering
\includegraphics[width=0.9\textwidth]{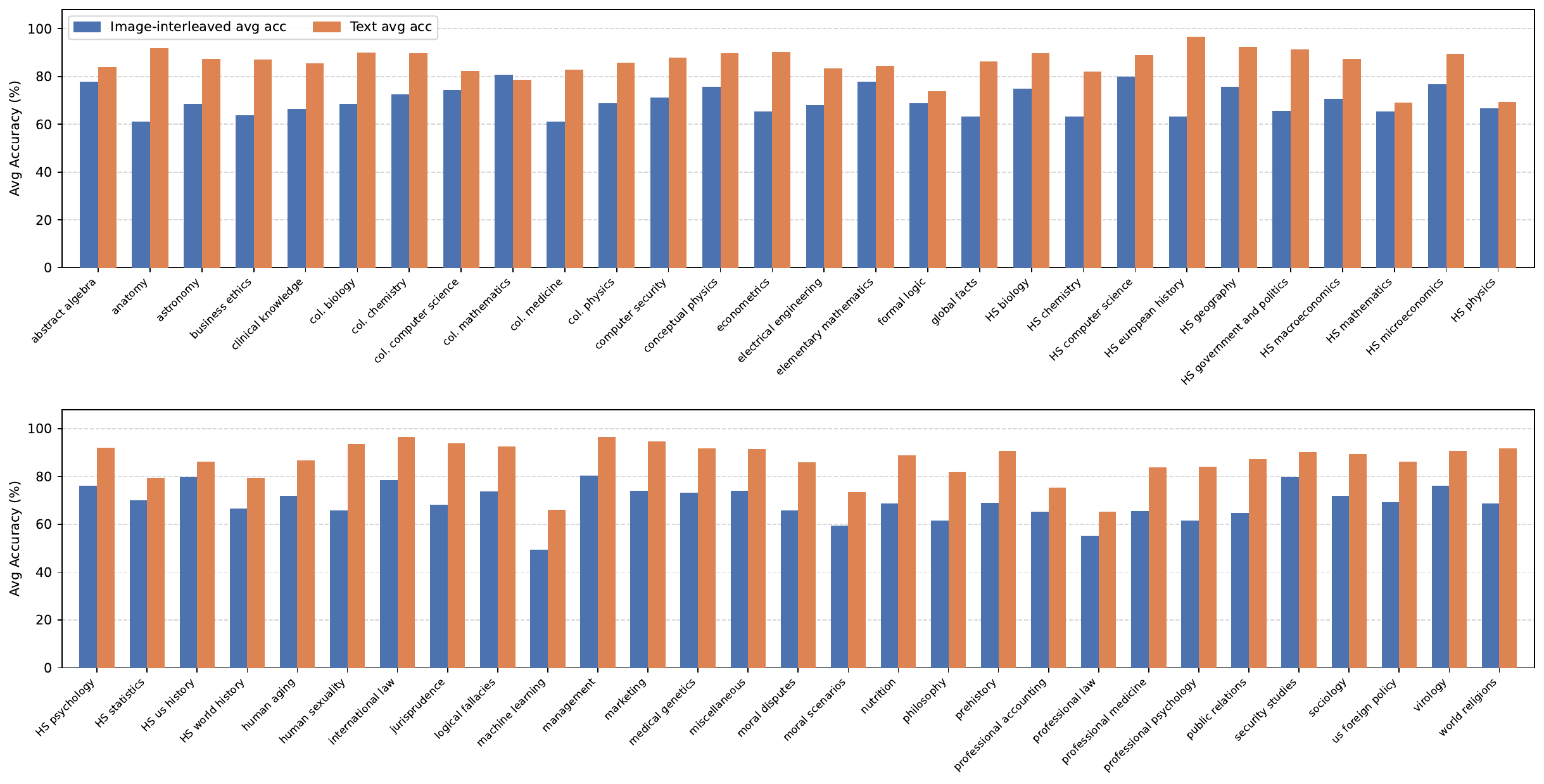}
\caption{Per-subject accuracy on MMLU comparing text-only and image-interleaved inputs.}
\label{fig:per_subject_bar}
\end{figure}

Figure~\ref{fig:per_subject_bar} presents per-subject results on \benchmark.
We observe a consistent modality gap across nearly all subjects, where image-interleaved performance is lower than text-only performance. 
The magnitude of the gap varies across domains: it is generally larger in knowledge-intensive and abstract subjects (e.g., history, law, and medicine), while relatively smaller in more structured or quantitative domains (e.g., mathematics and physics). 

One possible explanation is that, in quantitative domains such as mathematics, the core reasoning primarily depends on numerical values. While these values are less frequently replaced, even when they are presented visually, they are typically easy to interpret. In addition, although contextual concepts (e.g., "books" or "shelf") are important for understanding the problem, they do not directly affect the computation, which usually involves comparing or aggregating numerical values (e.g., identifying the valid range $2.3 \leq w \leq 3.2$ from the weights of the books on the shelf). As a result, models can still perform the required reasoning, leading to a smaller modality gap.

These results suggest that the modality gap is a broad and systematic phenomenon, while its magnitude varies across different subjects.

\begin{figure}
\centering
\includegraphics[width=0.4\textwidth]{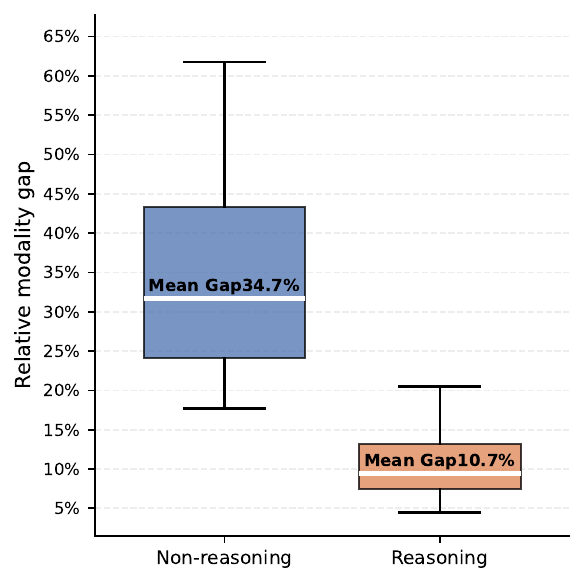}
\caption{Distribution of relative modality gap for reasoning and non-reasoning models.}
\label{fig:modality_gap_reasoning_relative}
\end{figure}

\subsection{Modality Gap Increases with the Number of Image Replacements}
\label{sec:gap_num_replacements}

\begin{figure}
\centering
\includegraphics[width=0.8\textwidth]{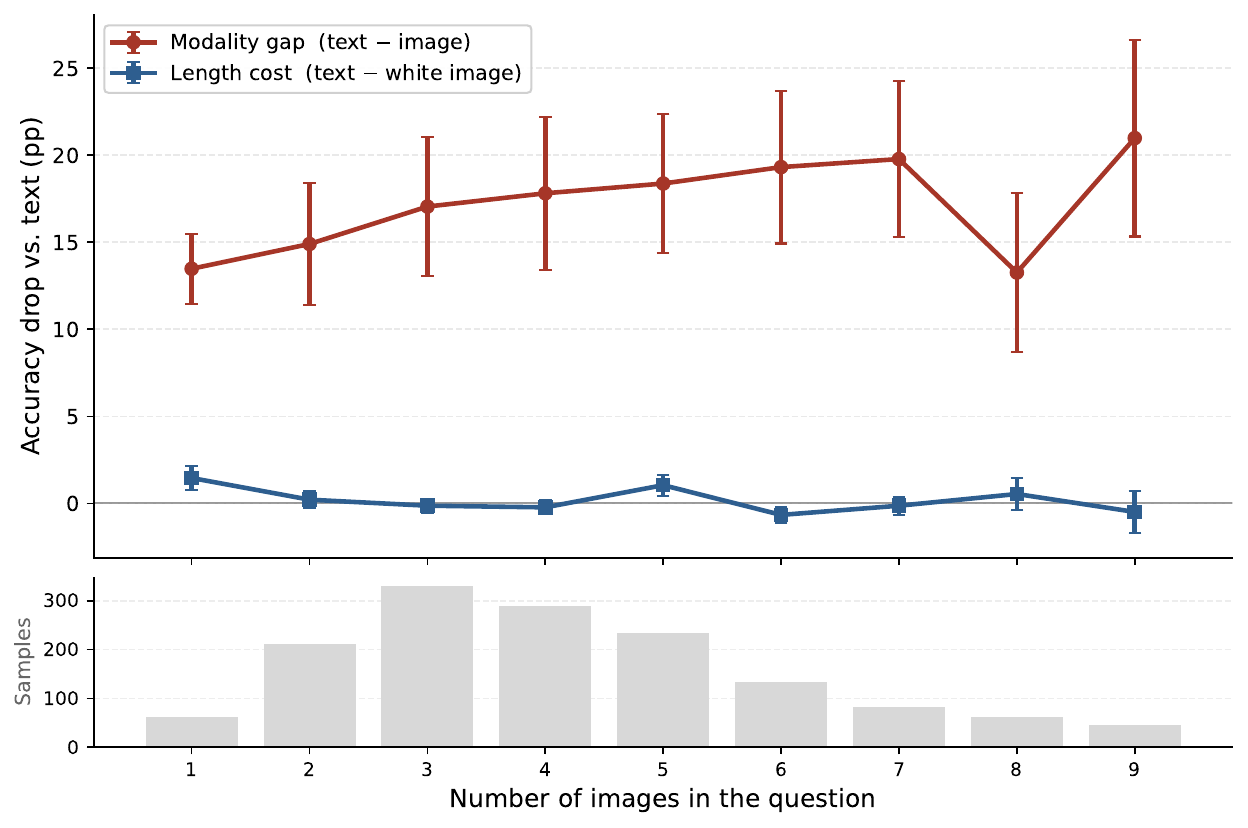}
\caption{\textbf{The modality gap generally increases with the number of image replacements, while the white-image gap remains near zero.}
We average results across nine models and group samples by the number of images. The natural-image gap (\text{Text} $-$ \text{Image}) generally increases as more textual concepts are replaced with images, whereas the white-image gap (\text{Text} $-$ \text{White Image}) remains close to zero, indicating that the trend cannot be explained by additional visual tokens alone. The bottom panel shows the number of samples in each group.}
\label{fig:modality_gap_vs_num_replacements}
\end{figure}

According to Figure~\ref{fig:modality_gap_vs_num_replacements}, we find that the modality gap generally increases as more text spans are replaced by images. Averaged across models, the gap rises from $13.5$ percentage points with one replacement to around $20$ percentage points with more replacements. This suggests that each additional image introduces extra visual grounding difficulty, making the text-to-image performance gap more pronounced.

To determine whether this trend is simply caused by processing more images, we perform the same analysis using the white-image images. In contrast to the natural-image gap, the white-image gap remains approximately flat around zero ($-0.7$ to $+1.5$ percentage points) as the number of images increases. This suggests that the growing modality gap is not caused by the number of images alone, but by moving more semantic information from text into the visual modality, which is exactly the modality gap we measures.

\subsection{Controlling for Visual-Input Confounds}
\label{app:modality_gap_controls}

\paragraph{Longer visual inputs.}
To isolate the effect of the additional visual tokens, we conduct a white-image ablation. For each \benchmark sample, we replace every image with a pure-white image of the same size, place all images at the beginning of the question, and restore the original text. This preserves the same number of visual tokens and results in an input length nearly identical to that of the corresponding \benchmark sample. 

As shown in Table~\ref{tab:white_image_control}, for all 9 models, the white-image accuracy stays within $\pm$1.0 pp of text-only accuracy, so the additional visual tokens contribute very little to the observed modality gap. These results show that the gap cannot be explained by the additional visual-token count or input length alone.

\begin{table}[t]
\centering
\caption{Detailed results for the white-image control.}
\label{tab:white_image_control}
\begin{tabular}{lccccc}
\toprule
Model & Text & White Image & Image &
\shortstack{Text $-$\\White Image} &
\shortstack{Text $-$ Image\\(Modality Gap)} \\
\midrule
Gemini-3-Flash & 0.925 & 0.935 & 0.884 & $-0.010$ & 0.041 \\
Claude-Opus-4.6 & 0.984 & 0.983 & 0.926 & 0.001 & 0.058 \\
GPT-5.1 & 0.980 & 0.981 & 0.903 & $-0.001$ & 0.077 \\
Qwen3-VL-4B-Thinking & 0.889 & 0.890 & 0.780 & $-0.001$ & 0.109 \\
GPT-4.1 & 0.925 & 0.923 & 0.761 & 0.002 & 0.164 \\
Qwen3-VL-4B & 0.707 & 0.714 & 0.523 & $-0.007$ & 0.184 \\
InternVL3-1B & 0.502 & 0.493 & 0.284 & 0.009 & 0.218 \\
InternVL3-78B & 0.943 & 0.938 & 0.685 & 0.005 & 0.258 \\
LLaVA-OV-7B & 0.710 & 0.702 & 0.301 & 0.008 & 0.409 \\
\bottomrule
\end{tabular}
\end{table}

\paragraph{Rendered-text-image.}
We also conduct a rendered-text-image baseline. We decompose the total modality gap into two components: the performance drop from text-only to rendered-text-image inputs, $(\text{text}-\text{rendered image})/(\text{text}-\text{real image})$, and the performance drop from rendered-text-image to natural concept-image inputs, $(\text{rendered image}-\text{real image})/(\text{text}-\text{real image})$.

As shown in Table~\ref{tab:rendered_text_control}, for seven of the nine models, 72--95\% of the gap lies in the second comparison: reading the phrase from a rendered image causes a substantially smaller performance drop of 0.4--7.0 percentage points. For the other two models, InternVL3-1B and LLaVA-OV, the gap comes mostly from the first comparison. However, both are among the weaker models in our evaluation pool, and this strong sensitivity to rendered-text-image is consistent with prior work~\citep{van2025same}.

\begin{table}
\centering
\caption{Detailed results for the rendered-text-image control.}
\label{tab:rendered_text_control}
\begin{tabular}{lccccc}
\toprule
Model & Text & Rendered & Image & Text$\rightarrow$Rendered & Rendered$\rightarrow$Image \\
\midrule
Gemini-3-Flash & 0.925 & 0.917 & 0.884 & 19.5\% & 80.5\% \\
Claude-Opus-4.6 & 0.984 & 0.976 & 0.926 & 13.8\% & 86.2\% \\
GPT-5.1 & 0.980 & 0.976 & 0.903 & 5.2\% & 94.8\% \\
Qwen3-VL-4B-Thinking & 0.889 & 0.876 & 0.780 & 11.9\% & 88.1\% \\
GPT-4.1 & 0.925 & 0.906 & 0.761 & 11.6\% & 88.4\% \\
Qwen3-VL-4B & 0.707 & 0.690 & 0.523 & 9.2\% & 90.8\% \\
InternVL3-1B & 0.502 & 0.297 & 0.284 & 94.0\% & 6.0\% \\
InternVL3-78B & 0.943 & 0.873 & 0.685 & 27.1\% & 72.9\% \\
LLaVA-OV-7B & 0.710 & 0.315 & 0.301 & 96.6\% & 3.4\% \\
\bottomrule
\end{tabular}
\end{table}

\subsection{Detailed Numerical Results}
\label{app:numeric}
We report detailed numerical results for \benchmark in Table~\ref{tab:numerical_number}, and present a comparison with SEAM~\citep{tang2025seam} in Table~\ref{tab:seam_numerical_number}.

\begin{table}
\centering
\caption{Detailed numerical results for \benchmark.}
\label{tab:numerical_number}
\begin{tabular}{lccc}
\toprule
Model & Text & Image & Gap \\
\midrule
GPT-4.1 & 0.925 & 0.761 & +0.164 \\
GPT-4.1-mini & 0.892 & 0.686 & +0.206 \\
GPT-4o & 0.913 & 0.693 & +0.220 \\
GPT-4o-mini & 0.832 & 0.545 & +0.288 \\
GPT-5 & 0.952 & 0.856 & +0.096 \\
GPT-5.1 & 0.980 & 0.903 & +0.077 \\
GPT-5.2 & 0.972 & 0.887 & +0.085 \\
Gemini-2.5-Flash & 0.954 & 0.885 & +0.070 \\
Gemini-2.5-Flash-Lite & 0.904 & 0.766 & +0.138 \\
Gemini-2.5-Pro & 0.968 & 0.888 & +0.080 \\
Gemini-3-Flash & 0.925 & 0.884 & +0.042 \\
Gemini-3.1-Flash-Lite & 0.954 & 0.759 & +0.195 \\
Gemini-3.1-Pro & 0.976 & 0.914 & +0.062 \\
Claude-Haiku-4.5 & 0.951 & 0.833 & +0.118 \\
Claude-Opus-4.6 & 0.984 & 0.926 & +0.057 \\
Claude-Sonnet-4.6 & 0.979 & 0.906 & +0.073 \\
InternVL2-40B & 0.856 & 0.505 & +0.352 \\
InternVL2-8B & 0.767 & 0.294 & +0.474 \\
InternVL2-1B & 0.453 & 0.184 & +0.268 \\
InternVL2.5-78B & 0.920 & 0.600 & +0.320 \\
InternVL2.5-8B & 0.756 & 0.382 & +0.374 \\
InternVL2.5-1B & 0.460 & 0.259 & +0.202 \\
InternVL3-78B & 0.943 & 0.685 & +0.258 \\
InternVL3-8B & 0.788 & 0.511 & +0.276 \\
InternVL3-1B & 0.502 & 0.284 & +0.218 \\
Qwen2-VL-72B & 0.900 & 0.726 & +0.174 \\
Qwen2-VL-7B & 0.704 & 0.481 & +0.223 \\
Qwen2-VL-2B & 0.555 & 0.346 & +0.209 \\
Qwen2.5-VL-72B & 0.909 & 0.731 & +0.178 \\
Qwen2.5-VL-7B & 0.677 & 0.529 & +0.148 \\
Qwen2.5-VL-3B & 0.670 & 0.458 & +0.211 \\
Qwen3-VL-32B & 0.889 & 0.690 & +0.199 \\
Qwen3-VL-30B-A3B-Instruct & 0.852 & 0.618 & +0.234 \\
Qwen3-VL-8B & 0.782 & 0.571 & +0.211 \\
Qwen3-VL-8B-Thinking & 0.928 & 0.808 & +0.120 \\
Qwen3-VL-4B & 0.707 & 0.523 & +0.184 \\
Qwen3-VL-4B-Thinking & 0.889 & 0.780 & +0.109 \\
LLaVA-OV-72B & 0.909 & 0.451 & +0.458 \\
LLaVA-OV-7B & 0.710 & 0.301 & +0.409 \\
LLaVA-OV-0.5B & 0.409 & 0.247 & +0.162 \\
GLM-4.1V-9B-Thinking & 0.901 & 0.740 & +0.161 \\
GLM-4.6V-Flash & 0.932 & 0.803 & +0.129 \\
\bottomrule
\end{tabular}
\end{table}

\begin{table}[h]
\centering
\caption{Comparison of modality gap between SEAM and our evaluation on overlapping models.}
\label{tab:seam_numerical_number}
\begin{tabular}{lcccccc}
\toprule
Model & \multicolumn{2}{c}{SEAM} & \multicolumn{2}{c}{Ours} & \multicolumn{2}{c}{Gap} \\
\cmidrule(lr){2-3}\cmidrule(lr){4-5}\cmidrule(lr){6-7}
 & Text & Image & Text & Image & SEAM & Ours \\
\midrule
  GPT-4o & 0.635 & 0.482 & 0.913 & 0.693 & +0.153 & +0.220 \\
  GPT-4o-mini & 0.555 & 0.411 & 0.832 & 0.545 & +0.144 & +0.288 \\
  InternVL2.5-78B & 0.448 & 0.414 & 0.920 & 0.600 & +0.034 & +0.320 \\
  InternVL2.5-8B & 0.324 & 0.337 & 0.756 & 0.382 & -0.013 & +0.374 \\
  InternVL3-78B & 0.525 & 0.427 & 0.943 & 0.685 & +0.098 & +0.258 \\
  InternVL3-8B & 0.382 & 0.357 & 0.788 & 0.511 & +0.025 & +0.276 \\
  Qwen2.5-VL-72B & 0.547 & 0.475 & 0.909 & 0.731 & +0.072 & +0.178 \\
  Qwen2.5-VL-7B & 0.303 & 0.350 & 0.677 & 0.529 & -0.047 & +0.148 \\
\bottomrule
\end{tabular}
\end{table}

\section{Details for TokenSwap Training}

\subsection{Experimental Setup}
\label{app:experimental_setup}
\paragraph{Data Construction.}
We start from a large-scale pure text instruction tuning dataset, Magpie-Pro~\citep{magpie}. We follow the same procedure described in Section~\ref{subsec:pipeline}, except that we remove \textit{Concept Importance Filtering} and \textit{Caption-Guided Validation} to enable scalable data generation. Specifically, we apply visual concept substitution only to the human queries while keeping the assistant responses unchanged.

\begin{figure}
\centering
\includegraphics[width=0.9\textwidth]{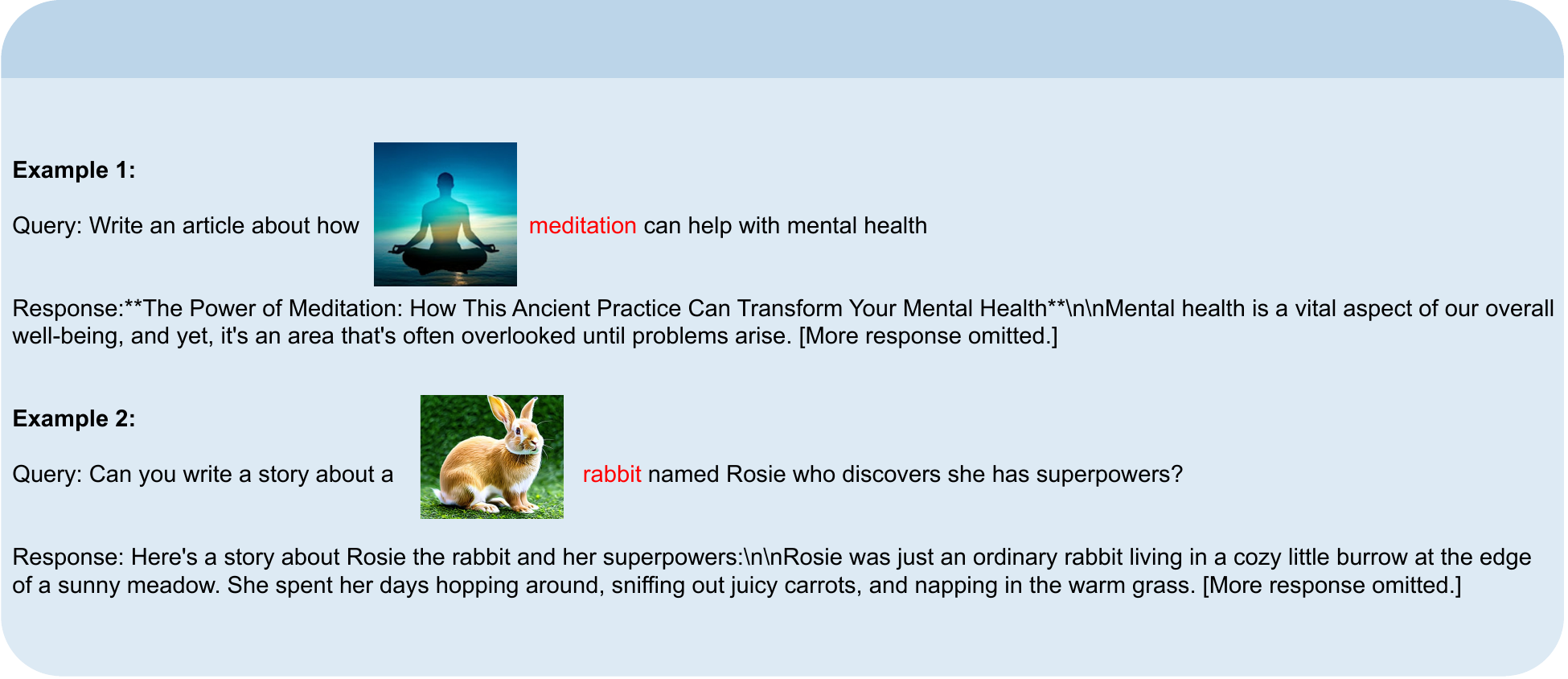}
\caption{\textbf{Applying TokenSwap to the Magpie-Pro dataset.}
We construct image-interleaved instructions by replacing key textual concepts in the original queries with semantically aligned images. The red text highlights the concepts that are replaced by images.}
\label{fig:magpie_example}
\end{figure}

This process yields 116,722 samples, with an average of 1.75 generated images per sample. Based on the constructed post-training data, we further derive a pre-training data variant. The pre-training data is nearly identical to the post-training data, except that the responses are removed. Example samples are provided in Figure~\ref{fig:magpie_example}.

We additionally explore an alternative data construction strategy using image retrieval. Specifically, we construct training data by retrieving images based on the LLaVA pre-training data~\citep{llava1.5}. Detailed implementation is provided in Appendix~\ref{app:training_retrieval}.

\paragraph{Experimental Setting.}
We compare two training paradigms: Text training and TokenSwap training. In text training, inputs remain purely textual, while TokenSwap training converts the same samples into image-interleaved inputs by replacing selected textual concepts with semantically aligned images. The two paradigms are constructed as \emph{paired settings}: both variants share the exact same samples and differ only in whether visual concepts are replaced, allowing us to evaluate whether TokenSwap training preserves text-only performance.

We use Qwen2-VL-7B as the base model\footnote{\url{https://huggingface.co/Qwen/Qwen2-VL-7B}} for all experiments and adopt the LLaVA post-training dataset\footnote{\url{https://huggingface.co/datasets/liuhaotian/LLaVA-Instruct-150K}} as the base post-training data. 
We choose Qwen2-VL-7B because it is one of the strongest models in our benchmark with publicly available pretrained weights, which are required for our controlled post-training experiments. 
Since Qwen does not release its post-training data, we use the LLaVA post-training dataset as a substitute. We study these paradigms across different training stages and image sources.

\textbf{Baseline.} We train the model using only the LLaVA post-training dataset.

\textbf{Continuous Pre-training.} We continue pre-training the base model using either pure text data or TokenSwap data. For TokenSwap training, we consider two variants based on the image source: generated images and retrieved images. All models are subsequently post-trained on the same LLaVA dataset as in the baseline.

\textbf{Post-Training.} For post-training, we directly perform visual instruction tuning on a mixture of baseline LLaVA data and our constructed data. Similar to pre-training, we consider both generated images and retrieved images.

\paragraph{Training Details.}
For post-training, we train the LLM and multimodal projector while keeping the vision encoder frozen. We train for one epoch with a learning rate of $2\times10^{-5}$ using a cosine learning-rate schedule and a warmup ratio of 0.03. We use a weight decay of 0, a maximum gradient norm of 1.0, and a per-device batch size of 4 with 4 gradient accumulation steps across 8 GPUs, resulting in an effective batch size of 128. Training uses bfloat16 precision, a maximum sequence length of 8192, and image resolutions ranging from 784 to 50,176 pixels. We use DeepSpeed ZeRO-3 with gradient checkpointing, and all experiments are conducted on 8 A100 GPUs.

\subsection{Generalization to a Second Training Backbone}
\label{app:second_backbone}

We replicate the post-training experiment on InternVL3-8B-Pretrained with the same training data (Gen variant) but at a smaller scale. Specifically, we use a random 12\% subsample of each training set.

\begin{table}[t]
\centering
\caption{TokenSwap training results on InternVL3-8B-Pretrained.}
\label{tab:internvl_training}
\begin{tabular}{lccc}
\toprule
Training Data & Text & Image & Gap \\
\midrule
Baseline & 0.741 & 0.438 & 0.303 \\
Text Training & 0.753 & 0.429 & 0.324 \\
TokenSwap Training & 0.748 & 0.518 & 0.230 \\
\bottomrule
\end{tabular}
\end{table}

As shown in Table~\ref{tab:internvl_training}, TokenSwap training reduces the gap from 0.303 to 0.230, and the improvement comes largely from the image-interleaved side (0.438 $\rightarrow$ 0.518), while text-only accuracy is almost unchanged (0.741 $\rightarrow$ 0.748). This reproduces our Qwen2-VL-7B finding on a second model family, showing that the mitigation is not specific to one backbone.

\subsection{Experimental Setup for OCR Task}
\label{app:ocr}

\paragraph{Data Construction.}

The IIIT 5K-Word dataset consists of image–text pairs, where each image contains a word and the corresponding text provides its transcription, with a standard split of 2,000 training and 3,000 test samples. Starting from the ground-truth text, we first use an LLM to construct natural language descriptions. We then apply TokenSwap by replacing the target word in the text with its corresponding image, resulting in image-interleaved samples.

\begin{figure}
\centering
\includegraphics[width=0.9\textwidth]{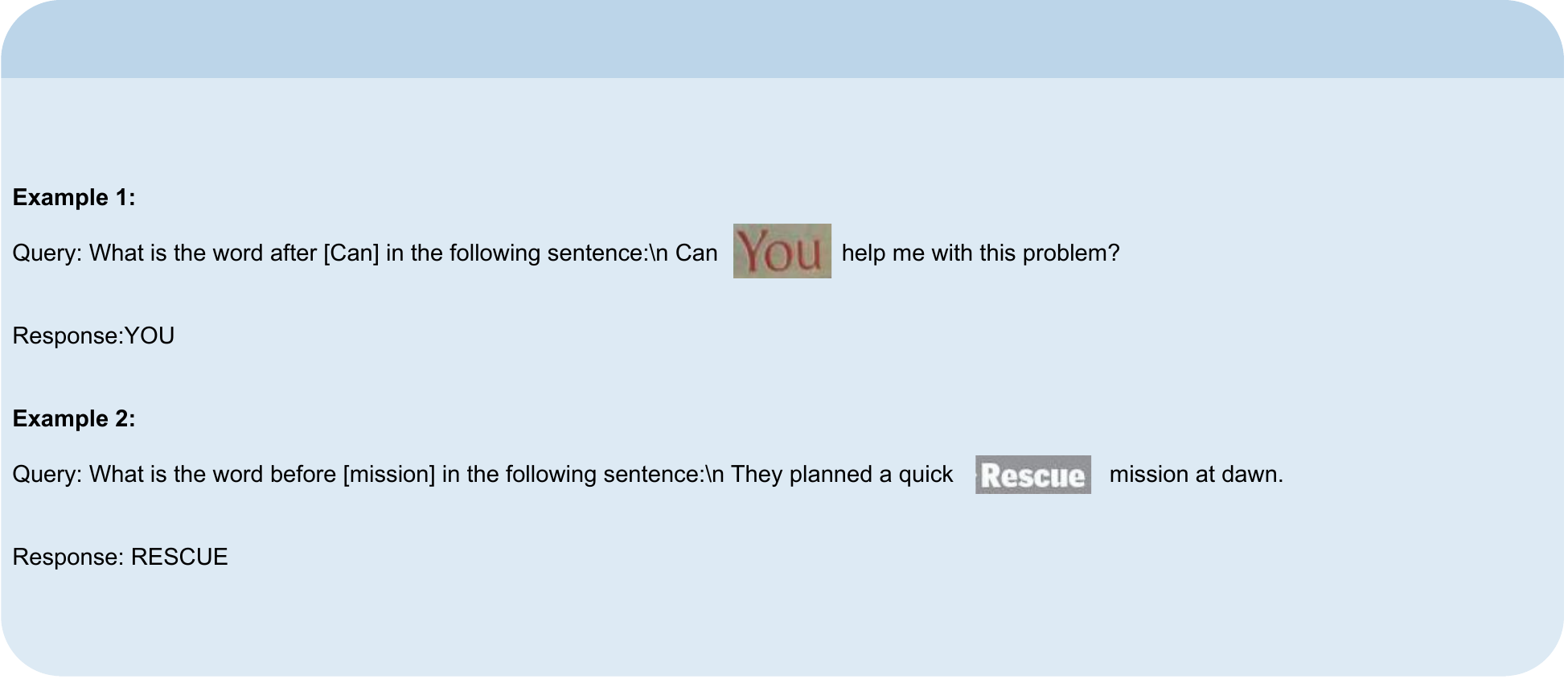}
\caption{\textbf{Applying TokenSwap to the IIIT 5K-Word dataset.}
We transform standard image–text pairs into image-interleaved instructions by replacing target words with images. The queries are generated from the ground-truth text using an LLM, where the OCR word is substituted with its image representation.}
\label{fig:iiit5k_example}
\end{figure}

To formulate the task as a conversational instruction, we construct user prompts that query the word shown in the image. For example, given an image–text pair where the image corresponds to the word ``RESCUE'', we create the prompt: ``What is the word before [mission] in the following sentence: They planned a quick <image> mission at dawn.'' Here, <image> represents the actual image of the target word, and the ground-truth answer is ``RESCUE''. We present more examples in Figure~\ref{fig:iiit5k_example}.

\paragraph{Experimental Setting.}

We follow the same experimental setup as in Section~\ref{app:experimental_setup}, using Qwen2-VL-7B as the base model. We augment the LLaVA post-training data (Baseline) with either our TokenSwap data (TokenSwap Training) or pure-text data (Text Training), and evaluate on the IIIT 5K-Word test split.

\begin{figure}
\centering
\includegraphics[width=0.85\textwidth]{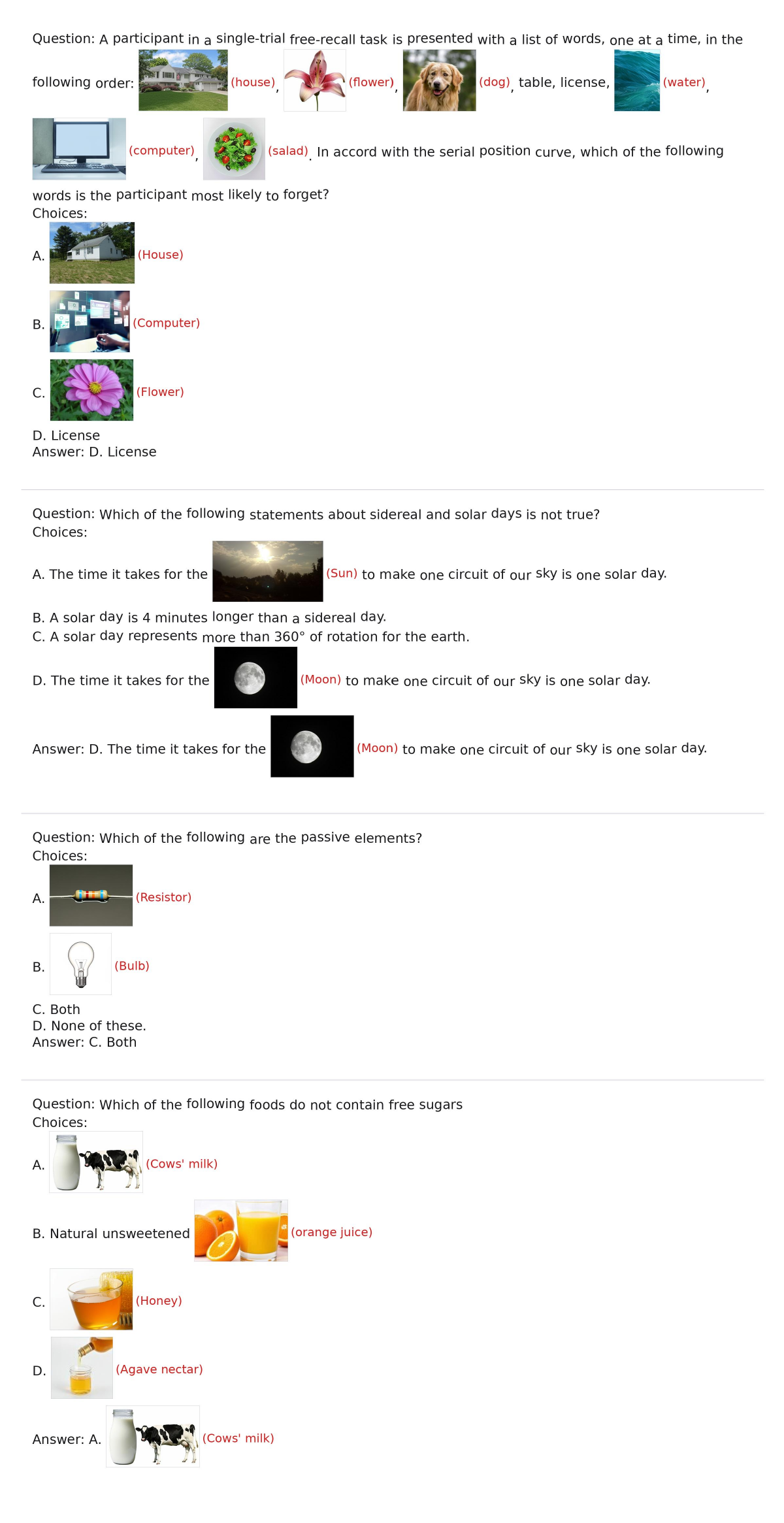}
\caption{Examples in \benchmark}
\label{fig:examples}
\end{figure}

\subsection{Details for Image Retrieval in TokenSwap Training}
\label{app:training_retrieval}

Unlike retrieving images from a large but noisy dataset such as DataComp-Small~\citep{datacomp}, as used in constructing \benchmark, we instead start from the curated image-text pairs used in LLaVA pre-training\footnote{\url{https://huggingface.co/datasets/liuhaotian/LLaVA-Instruct-150K/blob/main/llava_v1_5_mix665k.json}} to ensure higher data quality. For each image, we first perform a caption expansion step to enrich the textual description at multiple levels of granularity. Specifically, given an original caption (e.g., \textit{``the intel core processor''}), we expand it into a set of semantically equivalent captions ranging from coarse to fine-grained descriptions, such as \textit{``a computer processor''} and \textit{``a CPU chip''}.

Next, we use the expanded captions to retrieve candidate matches from the Magpie-Pro dataset~\citep{magpie} by performing exact string matching. To further ensure the quality of the retrieved pairs, we apply a filtering stage using \texttt{Qwen3-8B-Instruct} as a judge model. The model takes as input the full sentence together with the corresponding image, and is prompted to determine whether replacing the textual concept with the image preserves the original semantics. Only samples that are judged as valid replacements are retained for TokenSwap training. In total, this process yields 18,366 samples with 29,358 images. The prompt used for filtering is shown below.

\begin{promptbox}[Standard Prompt Template]
### Task
You are a strict data auditor. Your goal is to determine if an <image> can replace a word/phrase in the sentence UNAMBIGUOUSLY. For large-scale training, we require zero semantic drift. If you have any doubt, return false.

### Mandatory Decision Logic

1. **Part-of-Speech (POS) Match**:
   - The image MUST represent the word in its specific grammatical role in THIS sentence.
   - **RETURN FALSE IF**: The word is a verb but the image shows a noun (e.g., 'die' as in death vs. 'die' as a dice).

2. **Semantic Sense Alignment (Primary Defense)**:
   - Words often have multiple meanings. The image MUST match the exact sense used in the context.
   - **RETURN FALSE IF**:
     - Text means "human head" but image shows a "guitar amplifier head".
     - Text means "human nature" but image shows "forest/greenery".
     - Text means "tea time/meal" but image shows "tea leaves/packaging".

3. **Environmental & Adjective Consistency**:
   - The image must not contradict the sentence's setting or descriptors.
   - **RETURN FALSE IF**: Text describes a "hot desert" but image shows "wet sea sand"; text says "wrinkled" but image is "smooth".

4. **No Symbols, Icons, or Text**:
   - **RETURN FALSE IF**: The image is an icon, a diagram, a branded product, or contains significant printed text (like a book cover or a screenshot of words).

5. **Entity Specificity**:
   - **RETURN FALSE IF**: The image is a specific artifact (a toy, a statue, a specific brand) when the text refers to a generic living being or object.

---

### Calibration Examples:
- "The hot <image> (sand) scorched his feet." | Image: Ocean waves on sand. | Result: {{"i": 0, "word": "sand", "ok": false}}
- "The sun beat on his <image> (head)." | Image: A guitar amplifier. | Result: {{"i": 0, "word": "head", "ok": false}}
- "He sank down... again to <image> (die)." | Image: A gambling dice. | Result: {{"i": 0, "word": "die", "ok": false}}
- "But won't it be rather late after <image>?" (Word: tea) | Image: A branded tea box | Result: {{"i": 0, "word": "tea", "ok": false}}

---

### Data to Process:
Sentence: {text}
Replaced words: {word_list}

### Response (JSON ONLY):
{{
  "results": [
    {{
      "i": <int>,
      "word": "<string>",
      "ok": <boolean>
    }}
  ]
}}
\end{promptbox}

\subsection{Evaluation on Standard Vision-Language Benchmarks}
\label{app:vl_results}

We evaluate models trained with TokenSwap on standard vision-language benchmarks to assess whether the improved cross-modal consistency affects general multimodal performance. 

As shown in Table~\ref{tab:vl_results}, TokenSwap achieves comparable performance to the baseline across vision-language benchmarks, with no noticeable degradation. 

\begin{table}[h]
\centering
\caption{Evaluation on standard vision-language benchmarks (GQA, TextVQA, and MME-Peception). TokenSwap training achieves comparable performance to the baseline across all settings, indicating no noticeable degradation.}
\label{tab:vl_results}
\begin{tabular}{lrrr}
\toprule
 & GQA~\citep{gqa} & TextVQA~\citep{textvqa} & MME~\citep{mme} \\
\midrule
Baseline & 63.8 & 62.5 & 1542.4 \\
\midrule
pre-train + Generate & 63.5 & 61.7 & 1535.9 \\
pre-train + Retrieval & 63.7 & 61.9 & 1587.0 \\
post-train + Generate & 63.9 & 62.3 & 1560.0 \\
post-train + Retrieval & 63.8 & 62.0 & 1544.1 \\
\bottomrule
\end{tabular}
\end{table}

\subsection{Effect of Domain Mismatch on TokenSwap Training}
\label{app:domain_mismatch}

We study the impact of domain mismatch between TokenSwap training and evaluation across natural-image and OCR settings. 
For evaluation on the natural-image domain, we use image-interleaved performance on \benchmark.

As shown in Table~\ref{tab:domain_mismatch}, TokenSwap yields consistent improvements when training and evaluation domains are aligned. 
However, under domain mismatch, these gains are reduced or disappear, indicating limited cross-domain transferability.

\begin{table}[t]
  \caption{\textbf{TokenSwap training is sensitive to domain alignment}. 
  We consider two image domains: natural images and OCR. 
  We use image-interleaved performance on \benchmark to represent the natural-image domain, as it involves natural images rather than OCR-style text images. 
  TokenSwap improves performance only when the training and evaluation domains are aligned.}
  \label{tab:domain_mismatch}
  \centering
  \begin{tabular}{l c c}
    \toprule
    Training Domain & Natural Image-Interleaved Accuracy & OCR Accuracy \\
    \midrule
    No Train            & 0.4466 & 0.9003 \\
    Natural Image & 0.5132 & 0.9196 \\
    OCR Image     & 0.4393 & 0.9270 \\
    \bottomrule
  \end{tabular}
\end{table}

\section{Detailed Related Work}
\label{app:related_work}
\paragraph{Modality Gap in Vision--Language Models.}
The notion of \emph{modality gap} was originally studied in contrastive vision--language models~\citep{jia2021scaling,cherti2023reproducible,sun2023eva,zhai2023sigmoid,tschannen2025siglip}, particularly CLIP~\citep{clip}. 
Prior work shows that, even with a shared embedding space, image and text representations may remain separated, leading to a representation-level discrepancy between modalities~\citep{liang2022mind,shi2023understanding}. 
Subsequent studies further analyze this phenomenon from multiple perspectives, linking it to factors such as object bias and information imbalance~\citep{schrodi2024two}, exploring how it can be explained or mitigated~\citep{eslami2024mitigate,yaras2024explaining}, and even arguing that it arises from contrastive learning itself rather than modality differences~\citep{fahim2024s}.
However, such representation-level definitions do not naturally extend to multimodal large language models (MLLMs), which are trained via next-token prediction rather than explicit cross-modal alignment. In this work, we instead study the modality gap from a generative perspective, focusing on discrepancies in model predictions under semantically equivalent inputs.

\paragraph{Cross-Modal Consistency in MLLMs.}
A closely related line of work studies whether multimodal large language models behave consistently when the same semantic content is presented in different modalities~\citep{chen2024omnixr,yue2025mmmu,alonso2025vision}.
Early work studies cross-modal consistency by converting text into images via text rendering and evaluating MLLMs on semantically equivalent text--image pairs, revealing noticeable inconsistencies in model predictions~\citep{zhang2023lost,zhang2024cross}.
\citet{van2025same} further show that visual factors such as resolution, font, and color can affect cross-modal consistency, with models often achieving higher accuracy under certain color conditions. 
Besides rendered-text images, \citet{tang2025seam} construct semantically equivalent comparisons in controlled domains with standardized textual and visual notations (e.g., chess, chemistry, music, and graphs), and observes notable cross-modal inconsistencies.
XModBench~\citep{wang2025xmodbench} further extends cross-modal consistency evaluation to tri-modal settings involving text, vision, and audio.



\end{document}